\documentclass[lettersize,journal]{IEEEtran}
\usepackage{amsmath,amsfonts}
\usepackage{algorithmic}
\usepackage{algorithm}
\usepackage{array}
\usepackage[caption=false,font=normalsize,labelfont=sf,textfont=sf]{subfig}
\usepackage{textcomp}
\usepackage{stfloats}
\usepackage{url}
\usepackage{verbatim}
\usepackage{graphicx}
\usepackage{cite}
\usepackage{booktabs}
\usepackage{pifont}
\usepackage[table]{xcolor}
\usepackage{multirow}
\definecolor{gainGreen}{RGB}{0,120,0}

\usepackage{amsmath}
\definecolor{healthblue}{RGB}{223,236,248}
\definecolor{bikepurple}{RGB}{239,230,246}
\definecolor{cookingyellow}{RGB}{247,242,220}
\definecolor{deltagreen}{RGB}{232,250,244}
\definecolor{tokenblue}{RGB}{179,200,233}
\definecolor{jepagrey}{gray}{0.90}

\begin{document}

\title{From Where to How: Continuous 4D Interaction Forecasting from Egocentric Video}



\author{Qiaohui Chu,
        Haoyu Zhang,
        Meng Liu,~\IEEEmembership{Member,~IEEE},
        Haoxiang Shi,\\
        Dongmei Jiang,
        and Liqiang Nie,~\IEEEmembership{Senior Member,~IEEE}%
\thanks{Qiaohui Chu, Haoyu Zhang and Haoxiang Shi are with the School of Computer Science and
Technology, Harbin Institute of Technology (Shenzhen), Shenzhen 518055,
China, and Pengcheng Laboratory, Shenzhen 518000, China
(e-mail: qiaohuichu8599@gmail.com; zhang.hy.2019@gmail.com; shihaoxiang1999@gmail.com).}%
\thanks{Meng Liu is with the School of Software,
Shandong University, Jinan, 250101, China (e-mail: mengliu.sdu@gmail.com).}%
\thanks{Dongmei Jiang is with Pengcheng Laboratory, Shenzhen 518000, China
(e-mail: jiangdm@pcl.ac.cn).}%
\thanks{Liqiang Nie is with the School of Computer Science and Technology,
Harbin Institute of Technology (Shenzhen), Shenzhen 518055, China
(e-mail: nieliqiang@gmail.com).}%
\thanks{Corresponding authors: Meng Liu and Liqiang Nie.}%
}

\markboth{Journal of \LaTeX\ Class Files,~Vol.~14, No.~8, August~2021}%
{Shell \MakeLowercase{\textit{et al.}}: A Sample Article Using IEEEtran.cls for IEEE Journals}

\maketitle

\begin{abstract}
Egocentric 4D interaction forecasting aims to anticipate both \emph{where} future interactions will occur in 3D and \emph{how} the human body will move to realize them, providing an important capability for assistive robotics and human-computer interaction.
Existing methods struggle to translate semantic understanding into precise continuous 3D localization and to balance motion diversity with structural consistency in pose forecasting. More fundamentally, these tasks are often modeled separately, leaving the continuous geometric and temporal correspondence between interaction locations and body motion insufficiently captured.
To address these challenges, we introduce \emph{Coherent4D}, a large-scale egocentric dataset for continuous 4D interaction forecasting, comprising approximately 233K samples across three domains. Each sample pairs a sequence of future 3D interaction locations with corresponding full-body poses, aligned in time and expressed in a shared coordinate system. We also provide evaluation metrics in continuous space. Building on this formulation, we propose \emph{HIGFlow}, a Hand Interaction Guided Residual Flow framework that models forecasting as a cascaded where-to-how process. HIGFlow first forecasts continuous future interaction locations by combining semantic grounding with short-horizon visual dynamics, and then uses the predicted location sequence to condition a deterministic motion anchor and residual Flow Matching for diverse yet structurally consistent full-body motion forecasting. Extensive experiments across all three domains demonstrate consistent improvements over representative baselines on both location and pose forecasting, while ablations validate the contributions of the proposed components. The project page is available at \url{https://corrineqiu.github.io/from-where-to-how/}.
\end{abstract}


\section{Introduction}
\IEEEPARstart{E}{gocentric} 4D interaction forecasting aims to jointly anticipate where future interactions will occur in 3D space and how the human body will move to execute them. This capability is fundamental to proactive embodied intelligence,
where effective anticipation, planning, and assistance require a temporally continuous and geometrically consistent understanding of future locations and their associated body motions.
It can support applications such as task assistance~\cite{li2025satori}, risk warning~\cite{pei2025attentionar}, and human-robot collaboration~\cite{noormohammadi2025lead}.

Despite the inherently coupled nature of interaction location and body motion, existing research has largely approached them as two separate forecasting problems.
For \textbf{interaction location forecasting}, existing methods predict the spatial targets or trajectories of upcoming interactions from egocentric observations, including 
continuous hand motion regression using temporal or state-space models~\cite{bao2023uncertainty}, generative modeling of multiple plausible hand trajectories~\cite{hatano2025invisible}, and semantic or language-guided forecasting for object grounding and procedural reasoning~\cite{liu2025sfhand,chen2025flowing,seminara2026task,liu2026goal}.
In parallel, \textbf{full-body pose forecasting} predicts temporally evolving human poses from motion history and contextual cues.
Existing methods leverage pose history, egocentric observations, scene context, robot-view cues, or map-aware representations~\cite{zheng2022gimo,avogaro2024exploring,jiang2024map,liu2022investigating,shu2021spatiotemporal}, while stochastic diffusion- or flow-based methods model multiple feasible future motions~\cite{wei2023human,chen2023humanmac,lipman2022flow} and skeleton-aware approaches incorporate structural or kinematic priors to preserve valid body configurations~\cite{curreli2025nonisotropic}.
Together, these two lines of research provide complementary capabilities for predicting \emph{where} future interactions occur and \emph{how} the body moves to realize them, but do not explicitly model their temporal and geometric correspondence.
\IEEEpubidadjcol

FIction~\cite{ashutosh2025fiction} represents an early effort to bridge this gap by connecting interaction localization and pose forecasting within a unified dataset and task formulation. It represents future interaction locations as discrete voxel occupancy and predicts poses conditioned on candidate locations, thereby introducing an explicit dependency between the two predictions. 
However, interaction locations and full-body poses are not modeled as continuous, temporally aligned sequences in a shared metric space. As a result, the stepwise correspondence between the evolving interaction target and the body motion that realizes it remains unresolved.

\begin{figure}[!t]
\centering
\includegraphics[width=\linewidth]{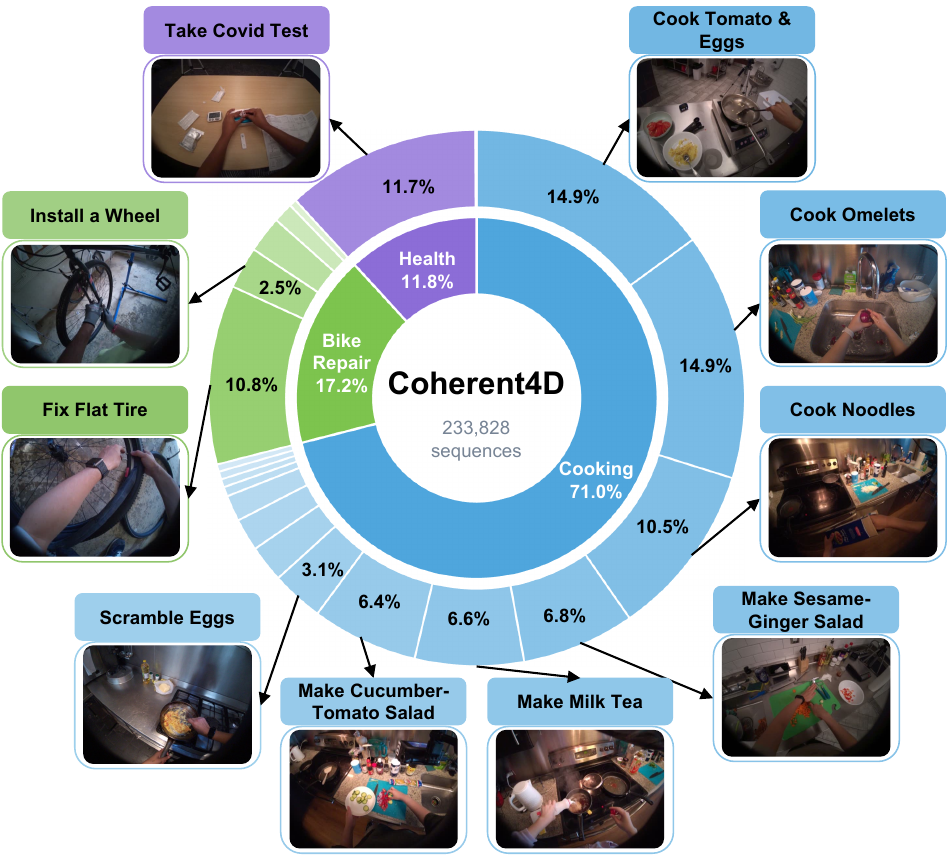}
\caption{Task distribution of Coherent4D. Sequences are associated with the corresponding task annotations to summarize the procedural coverage of the dataset.}
\label{fig:task_distribution}
\end{figure}

This limitation exposes a more fundamental gap: existing studies have not yet established a continuous and jointly grounded 4D forecasting formulation that preserves the temporal and geometric correspondence between future interaction locations and full-body motion. Addressing this gap involves four key challenges:
\textbf{\ding{182} Decoupled task formulation.}
Even when location and motion cues are jointly available, existing predictors typically treat interaction localization and pose forecasting as independent or loosely connected objectives. Consequently, continuous predictions of interaction locations are not explicitly propagated as geometric constraints for subsequent motion forecasting, limiting coordination between \emph{where} an interaction develops and \emph{how} it is physically realized. 
\textbf{\ding{183} Spatiotemporal pairing gap.}
Existing datasets rarely provide ordered, continuous 3D hand interaction locations paired with full-body poses at matched future timestamps in a shared metric coordinate system. 
Without such temporally synchronized and spatially co-registered targets, models cannot directly learn or evaluate how interaction locations and body motion co-evolve over time. 
\textbf{\ding{184} Semantic-dynamic localization gap.} Accurate location forecasting requires both task-level semantic grounding and precise continuous 3D localization over future time steps. However, representations from vision-language models (VLMs) are primarily optimized for semantic reasoning rather than metric coordinate regression and provide limited modeling of short-horizon visual dynamics. 
This mismatch can produce semantically plausible yet spatially inaccurate location predictions.
\textbf{\ding{185} Pose diversity-structure trade-off.} Deterministic pose forecasting tends to preserve structural consistency but suppresses alternative feasible executions, while stochastic generation captures multiple feasible futures but may compromise skeletal structure and joint consistency. Effective forecasting therefore requires motion diversity to be modeled without sacrificing structural plausibility.

To address these challenges, we introduce \textbf{Coherent4D}, a large-scale egocentric dataset for continuous 4D interaction forecasting. As shown in Fig.~\ref{fig:task_distribution}, Coherent4D contains approximately 233K samples spanning Cooking, Health, and Bike Repair. Unlike existing datasets that provide location and motion supervision separately or through discrete spatial representations, each sample in Coherent4D contains two temporally synchronized targets in a shared 3D coordinate system: an ordered sequence of continuous future interaction locations describing \emph{where} the interaction evolves over time, 
and a corresponding full-body pose sequence describing \emph{how} the body realizes it. 
This paired formulation establishes explicit temporal and geometric correspondence between interaction localization and motion realization.
We further introduce continuous-space evaluation metrics to assess both forecasts at corresponding future time steps.
Based on this coupled formulation, we propose \textbf{HIGFlow}, a cascaded framework for continuous 4D interaction forecasting that follows a structured \emph{where-to-how} process. 
HIGFlow first forecasts the continuous spatial progression of future locations and then uses these forecasts to condition the corresponding full-body motion, thereby explicitly coupling interaction localization with motion realization. 
For location forecasting, we develop \textbf{Semantic-Dynamic Location Forecasting}, which decouples task-level semantic grounding from continuous metric localization while incorporating short-horizon latent visual dynamics to improve future location prediction. 
For full-body pose forecasting, we introduce \textbf{Hand-Conditioned Residual Flow Matching}, which first constructs a deterministic motion anchor conditioned on the predicted future interaction locations and then models bounded stochastic residuals around this anchor. This design combines a geometrically grounded and structurally stable motion estimate with stochastic variation, enabling diverse future pose predictions while preserving skeletal structure and joint consistency. 
Extensive experiments validate the soundness of the Coherent4D dataset and the effectiveness of HIGFlow in jointly modeling interaction localization and full-body motion realization.

Our main contributions are summarized as follows:
\begin{itemize}

\item 
We introduce \textbf{Coherent4D}, a large-scale egocentric dataset with approximately 233K samples, pairing ordered future 3D interaction locations with temporally aligned full-body poses in a shared coordinate system.

\item 
We propose \textbf{HIGFlow}, a cascaded \emph{where-to-how} framework that uses predicted interaction locations as geometric conditions for full-body pose forecasting, explicitly coupling interaction localization with motion realization.

\item 
We develop \textbf{Semantic-Dynamic Location Forecasting} for accurate continuous interaction localization and \textbf{Hand-Conditioned Residual Flow Matching} for diverse yet structurally consistent full-body motion forecasting.

\item 
Extensive experiments demonstrate the effectiveness of Coherent4D and HIGFlow, as well as the benefit of jointly modeling future interaction locations and full-body motion.

\end{itemize}

\section{Related Work}

\begin{figure*}[!t]
\centering
\includegraphics[width=\textwidth]
{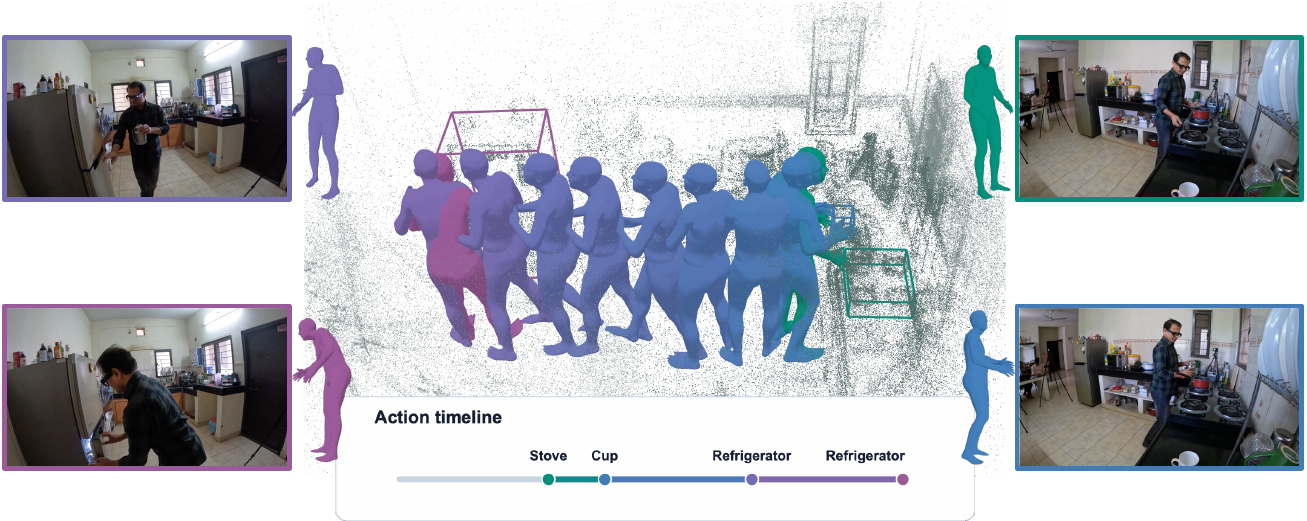}
\caption{Representative subset of interaction events from a Coherent4D sample. Synchronized exocentric source frames on both sides provide visual references for the selected events. The center shows the scene geometry, object bounding boxes, and temporally ordered SMPL body states in the shared sample-local coordinate frame. The action timeline summarizes the corresponding interaction objects, while colors associate each event with its rendered body state and reference frame.}
\label{fig:coherent4d_sample}
\end{figure*}

\subsection{Datasets for 4D Interaction Forecasting}

Large-scale egocentric datasets such as EPIC-KITCHENS and Ego4D provide broad supervision for action understanding and anticipation~\cite{damen2020epic,grauman2022ego4d}, but do not pair ordered continuous 3D interaction locations with temporally synchronized full-body poses.
Hand-centric datasets support forecasting future hand motion or interaction targets from egocentric observations. EgoPAT3D~\cite{li2022egocentric} and EgoPAT3Dv2~\cite{fang2024egopat3dv2} focus on 3D action target forecasting, while EgoHandTrajPred, introduced with USST~\cite{bao2023uncertainty}, provides ordered annotations for future 3D hand trajectory prediction. More recent datasets incorporate richer semantic and state information: EgoH4~\cite{hatano2025invisible} supports forecasting future 3D trajectories and poses of both hands, EgoHaFL~\cite{liu2025sfhand} provides language-guided annotations of future hand states, poses, and trajectories, and EgoMAN~\cite{chen2025flowing} supports 6-DoF hand trajectory forecasting with location awareness, using structured semantic, spatial, and motion cues. In contrast, pose-centric datasets focus on future full-body motion. MoGaze~\cite{kratzer2020mogaze} and GIMO~\cite{zheng2022gimo} incorporate workspace geometry, scene scans, egocentric observations, and gaze, while HARPER~\cite{avogaro2024exploring} and Real-IM~\cite{jiang2024map} extend pose forecasting to robot-centric and map-aware settings. However, hand-centric datasets generally lack temporally aligned full-body pose sequences, whereas pose-centric datasets rarely annotate future interaction moments. Consequently, existing datasets cannot directly associate future interaction locations with the corresponding full-body motion or evaluate whether the predicted body reaches the intended interaction target at the correct time.

FIction~\cite{ashutosh2025fiction} is the closest prior dataset linking future interaction localization with body pose forecasting. However, it represents locations as voxel occupancy and conditions pose prediction on individual candidate locations, rather than providing continuous, temporally aligned sequences of interaction locations and full-body poses. Coherent4D fills this gap by pairing ordered 3D interaction locations with synchronized full-body poses parameterized by the Skinned Multi-Person Linear (SMPL)~\cite{loper2023smpl} model in a shared coordinate system, enabling joint evaluation of interaction localization and motion execution.

\subsection{3D Interaction Location Forecasting}

3D interaction location forecasting predicts ordered metric hand interaction locations from egocentric observations, specifying \emph{where} upcoming hand-environment interactions will occur. Related egocentric studies address long-term anticipation and next active object prediction~\cite{qi2024uncertainty,peirone2026hier,mur2026integrating}. Early hand trajectory forecasting methods mainly operate in the 2D image space. OCT predicts future 2D hand trajectories with interaction hotspots, while Diff-IP2D and MADiff use diffusion- or dynamics-aware modeling to capture future uncertainty and ego-motion effects~\cite{liu2022joint,ma2025diff,ma2026madiff}. Recent works extend hand forecasting to metric 3D space: USST establishes egocentric 3D hand trajectory forecasting from RGB observations, MMTwin extends the MADiff-style diffusion paradigm to multimodal 3D hand trajectory forecasting, and Uni-Hand further unifies 2D/3D hand waypoint forecasting with richer hand motion and interaction targets~\cite{bao2023uncertainty,ma2025novel,ma2026uni}. Although these methods advance hand motion forecasting from image-space trajectories to metric 3D waypoints, they still primarily model future hand motion itself, without sufficiently integrating task-level semantic grounding, continuous metric regression, and short-horizon visual dynamics for ordered future interaction location forecasting.

Semantic and VLM-assisted methods introduce object-centric grounding, action semantics, and procedural context~\cite{zeng2023x,tian2026ego,Chen2025MotionLLM,Liu2026STKAD}. HandsOnVLM shows that directly serializing coordinates as language is insufficient, and instead uses dedicated hand tokens with a trajectory decoder~\cite{Bao2025HandsOnVLM}. Predictive video models and latent world representations further capture short-horizon spatiotemporal dynamics~\cite{Chang2025STAU,Maes2026LeWorldModel}. However, semantic reasoning, continuous metric regression, and predictive visual dynamics are still rarely integrated in a unified interaction location predictor. To address this limitation, our Semantic-Dynamic Location Forecasting separates task-level VLM grounding from coordinate decoding and augments the future position representation with short-horizon visual dynamics, improving goal consistency while preserving deterministic continuous 3D regression.

\subsection{Full-Body Pose Forecasting}

Full-body pose forecasting predicts future global motion and articulated body configurations from observed pose histories and contextual cues~\cite{Fernando2025Remembering,Tang2026ContinualPrior}. Prior methods improve motion stability using gaze, workspace context, spatiotemporal anchors, global trajectories, or affordance cues. MoGaze incorporates gaze and scene context, STARS separates deterministic anchors from within-mode variation, T2P conditions local pose forecasting on predicted global trajectories, and GAP3DS uses gaze-informed affordance cues in 3D scenes~\cite{kratzer2020mogaze,Xu2022STARS,Jeong2024T2P,Yu2025GAP3DS}. These methods improve scene consistency and stable coarse motion, but deterministic or anchor-based forecasting often suppresses alternative feasible executions when the same interaction can be realized by different body motions.

Other methods improve motion diversity through latent variables, diffusion, flow matching, motion fields, skeleton-aware generation, or residual refinement~\cite{Zhu2024HumanMotionSurvey,Zhang2024MotionDiffuse,Yang2026LagrangianMotionFields}. SLD-HMP learns controllable semantic latent directions, BeLFusion and CoMusion improve diverse motion generation and history consistency, SkeletonDiffusion incorporates structural priors for anatomical plausibility, and PrediFlow refines coarse motion forecasts with Flow Matching residuals~\cite{Xu2024SLDHMP,barquero2023belfusion,Sun2024CoMusion,curreli2025nonisotropic,Tian2025PrediFlow}. While these methods enhance diversity or realism, noise-initialized generation can still introduce inefficient sampling, temporal instability, or implausible articulation when stochastic variation is not sufficiently constrained. Our Hand-Conditioned Residual Flow Matching addresses this diversity-stability trade-off by first establishing a hand-conditioned deterministic SMPL motion anchor and then modeling bounded stochastic residuals around it.

\section{Coherent4D Dataset}

Coherent4D is constructed from Ego-Exo4D~\cite{Grauman2024EgoExo4D} to support continuous 4D interaction forecasting from egocentric video. Here, continuous refers to both the temporal continuity of motion sequences and the use of continuous 3D coordinates rather than discrete spatial grids.
It pairs ordered 3D interaction locations with SMPL-parameterized full-body pose sequences at matched future timestamps in a shared sample-local coordinate frame, jointly capturing the spatial progression of future locations and the corresponding body motion. 
Each sample contains 30 egocentric frames uniformly sampled from a 30-second observation window, structured object and environment descriptors, histories of observed interaction locations and SMPL poses, and the corresponding future interaction location and pose targets.
Fig.~\ref{fig:coherent4d_sample} illustrates representative interaction events from a Coherent4D sample. 
Each event associates a continuous 3D interaction location with the corresponding SMPL body state at the same timestamp, explicitly preserving their temporal and spatial correspondence.


\subsection{Annotation Pipeline}
\begin{figure}[!t]
\centering
\includegraphics[width=\columnwidth,keepaspectratio]
{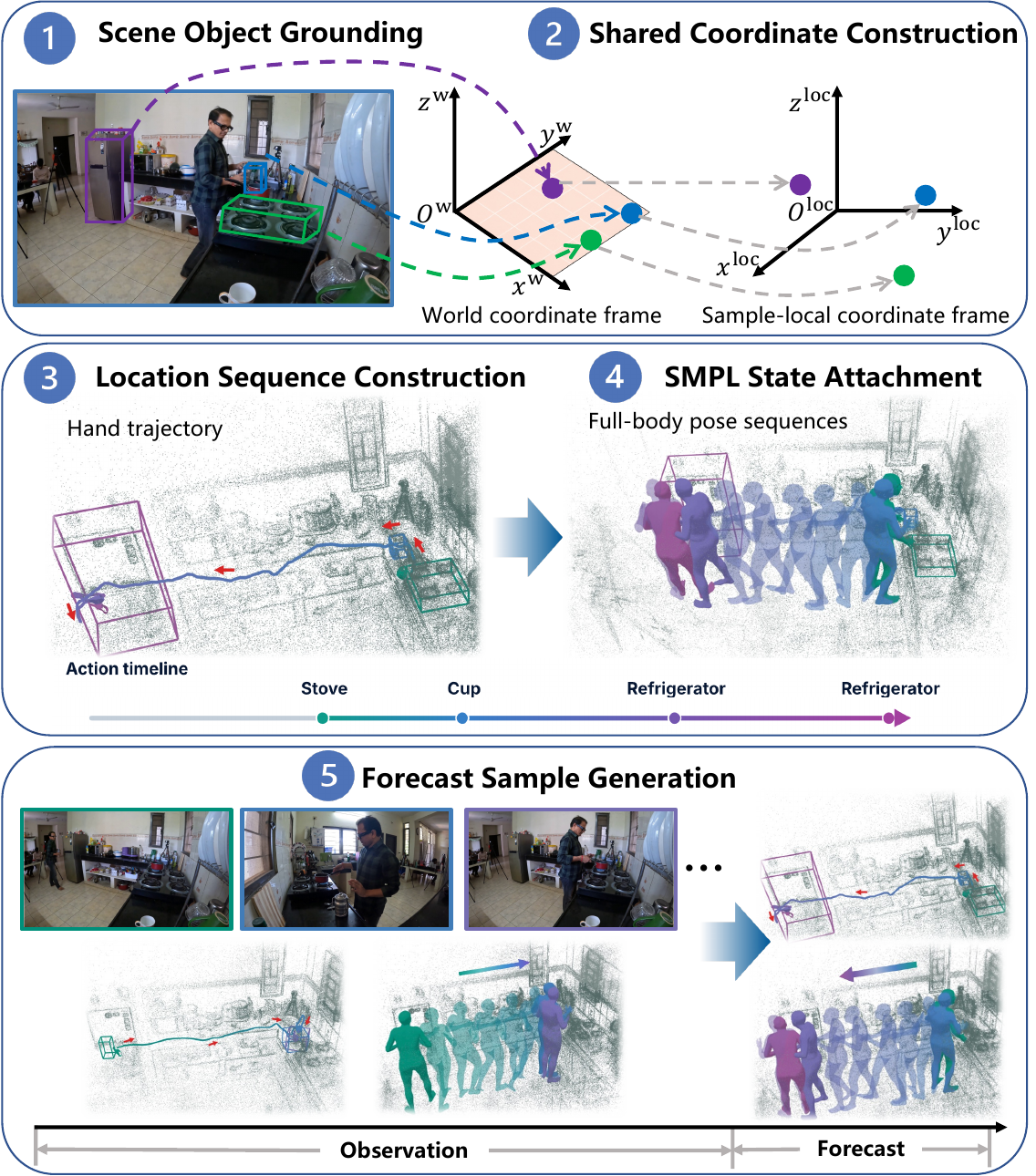}
\caption{\textbf{Overview of the Coherent4D data construction pipeline.} We (1) ground scene objects with semantic labels and 3D bounding boxes, (2) transform scene, location, and body annotations into a shared sample-local coordinate frame, (3) construct temporally ordered 3D interaction location sequences, (4) attach time-aligned SMPL states, and (5) organize them into forecasting samples with location-pose histories, future targets, and relative timestamps.}
\label{fig:annotation_pipeline}
\end{figure}

We construct Coherent4D from synchronized procedural takes in Ego-Exo4D~\cite{Grauman2024EgoExo4D}.
The Aria egocentric stream serves as the forecasting input, while synchronized exocentric views are used only for offline annotation construction and refinement. To establish temporally and geometrically coupled supervision, we organize the source annotations into ordered interaction location and SMPL pose sequences in a shared sample-local coordinate frame.
As illustrated in Fig.~\ref{fig:annotation_pipeline}, the pipeline consists of five stages: scene object grounding, shared coordinate construction, location sequence construction, SMPL state attachment, and forecast sample generation.

\subsubsection{\textbf{\textit{Scene Object Grounding}}}
We construct object-level semantic and geometric context for interaction location forecasting.
We apply Detic~\cite{zhou2022detecting} with the LVIS vocabulary~\cite{Gupta2019LVIS} to detect candidate objects in the egocentric frames. The detected regions are lifted into 3D using the Ego-Exo4D SLAM reconstruction and subsequently clustered to obtain oriented 3D bounding boxes.
For each object, we retain its semantic category together with continuous 3D bounding box attributes, including center, size, and orientation. The spatial attributes are subsequently expressed in the shared sample-local coordinate frame defined below, while the box dimensions retain their metric scale. This representation preserves both object semantics and continuous scene geometry without discretizing the environment into voxels.

\subsubsection{\textbf{\textit{Shared Coordinate Construction}}}
Geometric coupling between interaction location, body motion, and scene geometry requires all spatial quantities to be represented in a common coordinate frame. 
For each sample \(n\), we define a stable egocentric reference pose
\((\mathbf R_{n}^{\mathrm{ref}},\mathbf t_{n}^{\mathrm{ref}})\). Given a point \(\mathbf p^{\mathrm w}\) in the world coordinate frame, its representation in the shared sample-local coordinate frame is defined as:
\begin{equation}
\mathbf p^{\mathrm{loc}}_{n}
=
\left(\mathbf R_{n}^{\mathrm{ref}}\right)^{\top}
\left(
\mathbf p^{\mathrm w}
-
\mathbf t_{n}^{\mathrm{ref}}
\right).
\end{equation}
We apply this transformation to interaction locations, object centers, SMPL root translations, and 3D body joints. Global orientations, including the SMPL root orientation and object orientations, are rotated by the same reference rotation, whereas the 23 SMPL body joint rotations remain defined relative to their parent joints. For model input and supervision, positional quantities are divided by \(s = 5\,\mathrm{m}\) and then clipped to \([-1,1]\) for each coordinate. Division by \(s\) rescales the coordinates without changing the reference frame, whereas clipping limits values outside this range to the corresponding boundary. 

For model input and supervision, coordinates are normalized by dividing by \(s = 5\,\mathrm{m}\) and clipping to \([-1,1]\), which rescales their values without changing the reference frame. Root translations and translation residuals remain in meters for pose residual computation, clipping, and composition.

\subsubsection{\textbf{\textit{Location Sequence Construction}}}
We convert sparse hand-object interaction annotations into temporally ordered continuous 3D interaction location sequences. Following FIction~\cite{ashutosh2025fiction}, candidate interaction events are identified by combining narration timestamps, Llama 3-based object matching~\cite{grattafiori2024llama3}, and geometric consistency between the hand and the corresponding object. Each valid event retains its timestamp, interaction object when available, and continuous 3D interaction location recovered from the corresponding hand mesh. Interactions involving the right hand or both hands are mapped into a unified interaction stream. Temporally adjacent events with redundant spatial locations are merged according to their timestamps and normalized spatial displacement. The resulting sequence therefore contains distinct interaction locations while preserving their chronological order and continuous spatial evolution. 


\subsubsection{\textbf{\textit{SMPL State Attachment}}}
We then associate each interaction location with the full-body state at the corresponding timestamp. We reconstruct human motion using WHAM~\cite{Shin2024WHAM}, select the primary actor track, and align its SMPL trajectory with the Ego-Exo4D scene coordinate system. For every observed and future interaction timestamp, we retrieve the nearest valid SMPL state and transform its global components into the sample-local frame. 
Each SMPL state contains the root translation, root orientation, body joint rotations, and 3D joint positions. The root translation and 3D joints are thus spatially co-registered with the corresponding interaction location, while body joint rotations remain relative to their parent joints. This step yields temporally paired interaction location and full-body pose sequences for subsequent forecasting.


\begin{table}[!t]
\caption{Statistics of Coherent4D. Train, Val, Test, and Total report the numbers of forecasting samples. Targets denote valid future interaction locations, excluding padded steps. Obj. denotes interaction object labels, while the Total entry for Obj. reports the number of unique labels across the dataset.}
\label{tab}
\centering
\setlength{\tabcolsep}{1.5pt}
\begin{tabular}{@{}lcccccccc@{}}
\toprule
Domain & Horizon & Takes & Train & Val & Test & Total & Targets & Obj. \\
\midrule
Cooking & 10 & 331 & 136,079 & 15,435 & 14,527 & 166,041 & 1,337,589 & 428 \\
Health & 5 & 205 & 21,532 & 1,899 & 4,167 & 27,598 & 114,964 & 170 \\
Bike Repair & 4 & 251 & 35,987 & 3,150 & 1,052 & 40,189 & 141,633 & 197 \\
\midrule
Total & -- & 787 & 193,598 & 20,484 & 19,746 & 233,828 & 1,594,186 & 535 \\
\bottomrule
\end{tabular}%
\end{table}

\begin{figure}[!t]
\centering
\includegraphics[width=\linewidth]{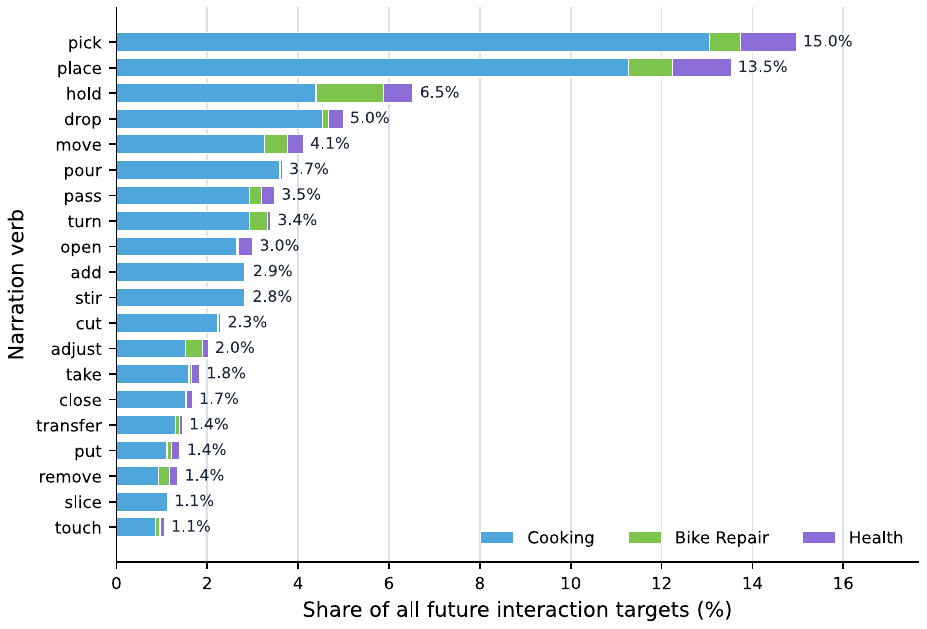}
\caption{Narration verb distribution over future interaction locations. Bars show the share of all future interaction targets for each normalized first narration verb, and colors indicate the contribution from each domain.}
\label{fig:verb_distribution}
\end{figure}

\subsubsection{\textbf{\textit{Forecast Sample Generation}}}
Finally, we convert the aligned annotation streams into forecasting samples while retaining the nonuniform timing of interaction events. The ordered sequences are partitioned according to the domain-specific forecasting horizons reported in Table~\ref{tab}. 
For each sample \(n\), we take the first future interaction timestamp as the time origin and denote it by \(t_{n,1}\).
The preceding 30 seconds constitute the observation window and provide the egocentric frames, location history, and pose history.
For each future step \(k\in\{1,\ldots,K\}\), we retain its absolute timestamp \(t_{n,k}\), relative timestamp
\(\Delta t_{n,k}=t_{n,k}-t_{n,1}\),
continuous interaction location, and temporally aligned SMPL state. The relative timestamps preserve the nonuniform intervals between interaction events, while a fixed number of target steps provides a consistent sequence interface for training and evaluation. Incomplete tail sequences are retained using padding and validity masks. Samples with invalid timestamp alignment, object grounding, coordinate transformation, interaction locations, or SMPL attachment are discarded. We split the dataset at the take level to prevent overlap of environments and procedural sequences across the training, validation, and test sets.


\subsection{Dataset Statistics}

\begin{figure*}[!t]
\centering
\includegraphics[width=\textwidth]{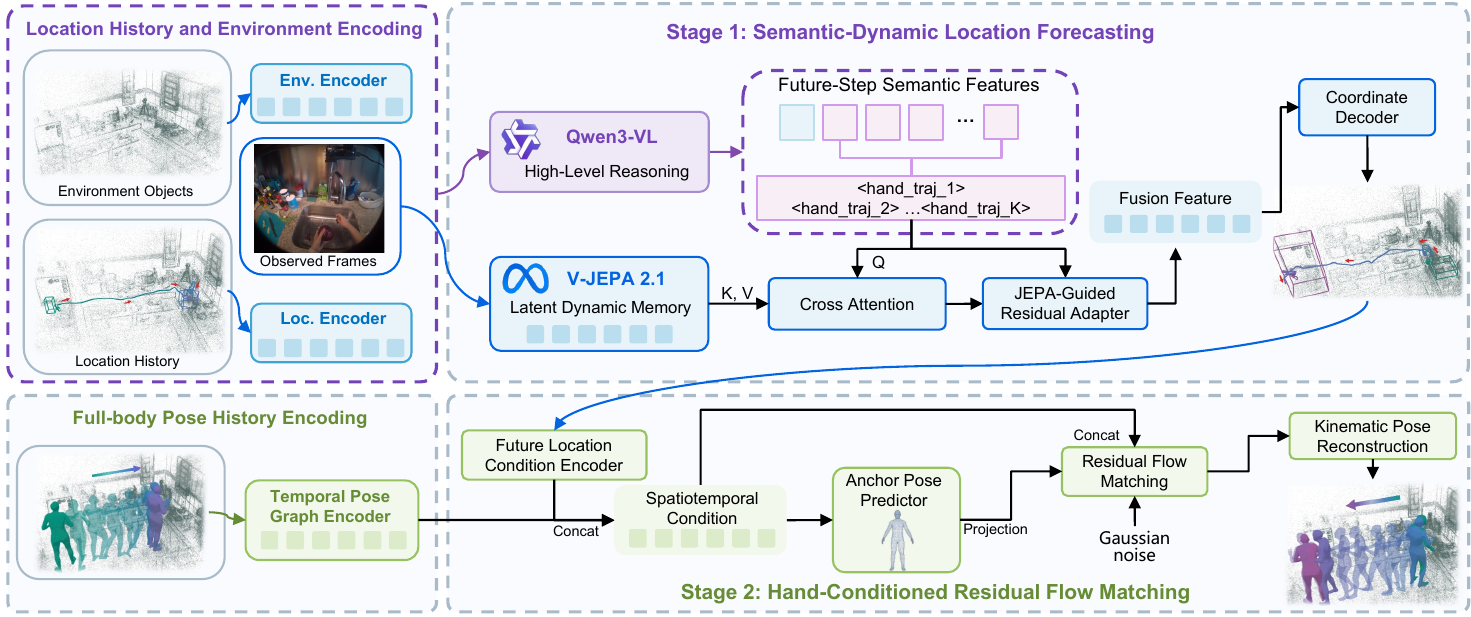}
\caption{Overview of HIGFlow. The first stage forecasts the locations of future interactions from egocentric context, and the second stage forecasts temporally aligned full-body poses conditioned on the predicted location sequence.}
\label{fig:framework}
\end{figure*}

As summarized in Table~\ref{tab}, Coherent4D contains 233,828 samples derived from 787 unique takes across three domains: Cooking, Health, and Bike Repair. The corresponding forecasting horizons are 10, 5, and 4 interaction steps, respectively. The dataset further covers 535 interaction object categories, providing diverse object-centric procedural activities across the three domains.


To characterize action-level diversity, we associate each valid future interaction target with its corresponding narration. 
All 1,594,186 valid future targets are aligned with narration annotations. Fig.~\ref{fig:verb_distribution} shows the normalized distribution of the first verb in each narration, covering common manipulation primitives such as pick, place, hold, drop, move, pour, pass, and turn. Cooking accounts for most high-frequency manipulation actions, while Bike Repair and Health contribute complementary domain-specific interaction patterns.



\subsection{Continuous Metrics}
Voxel-based accuracy measures whether a discrete spatial cell is correctly activated, but does not quantify continuous 3D localization or motion errors. We therefore introduce continuous-space metrics for both interaction location and full-body pose forecasting. For evaluation, normalized interaction locations are converted back to metric coordinates in the shared sample-local frame using the position scale \(s\). SMPL root translations and 3D joints are evaluated in the same metric frame. Positional and rotational errors are reported in millimeters and degrees, respectively, with lower values indicating better performance.

Let \(\mathcal D\) denote the evaluation set and \(K\) the forecasting horizon. For sample \(n\) and future step \(k\), let \(m_{n,k}\in\{0,1\}\) indicate whether the paired interaction location and pose targets are valid.


\subsubsection{\textbf{\textit{Interaction Location Forecasting Metrics}}}
To characterize overall sequence accuracy, upper-tail localization error, and endpoint accuracy, we report average displacement error (ADE), \(\mathrm{ADE}_{90}\), and final displacement error (FDE), respectively.
For sample \(n\) and future step \(k\), let \(\mathbf y_{n,k}\in\mathbb{R}^{3}\) and \(\hat{\mathbf y}_{n,k}\in\mathbb{R}^{3}\) denote the ground-truth and predicted interaction locations, respectively. The localization error at each future step is \(d^{\mathrm{loc}}_{n,k}
=
\left\|
\hat{\mathbf y}_{n,k}
-
\mathbf y_{n,k}
\right\|_2\).
We compute the ADE by first averaging over the valid future steps of each sample and then across samples:
\begin{equation}
\mathrm{ADE}
=
\frac{1}{|\mathcal D|}
\sum_{n\in\mathcal D}
\frac{
\sum_{k=1}^{K}
m_{n,k}d^{\mathrm{loc}}_{n,k}
}{
\sum_{k=1}^{K}m_{n,k}
}.
\end{equation}
We additionally report \(\mathrm{ADE}_{90}\), defined as the 90th percentile of the per-sample ADE values, to characterize upper-tail localization error.
FDE measures localization accuracy at the last valid future step. Let \(\kappa_n=\max\{k:m_{n,k}=1\}\) denote the last valid step of sample \(n\). FDE is then defined as:
\begin{equation}
\mathrm{FDE}
=
\frac{1}{|\mathcal D|}
\sum_{n\in\mathcal D}
d^{\mathrm{loc}}_{n,\kappa_n}.
\end{equation}

\subsubsection{\textbf{\textit{Full-body Pose Forecasting Metrics}}}
To assess absolute joint position accuracy, articulated pose accuracy after similarity alignment, global body displacement, and local body rotation, we report mean per-joint position error (MPJPE), Procrustes-aligned mean per-joint position error (PA-MPJPE), root translation error (Root Trans.), and body geodesic error (Body Geo.), respectively.
Following FIction~\cite{ashutosh2025fiction}, we evaluate each SMPL
state using the same set of \(J_{\mathrm{pos}}=19\) body joints.
Let \(\mathbf P_{n,k},\hat{\mathbf P}_{n,k}
\in\mathbb{R}^{3\times J_{\mathrm{pos}}}\)
denote the ground-truth and predicted joint coordinate matrices for
sample \(n\) at future step \(k\), with
\(\mathbf p_{n,k,j},\hat{\mathbf p}_{n,k,j}\in\mathbb{R}^{3}\)
denoting the corresponding \(j\)-th joint coordinates. 
For compact notation, let
\(
N_{\mathrm{valid}}
=
\sum_{n\in\mathcal D}
\sum_{k=1}^{K}
m_{n,k}
\)
denote the total number of valid future steps.

MPJPE measures the average Euclidean distance between corresponding predicted and ground-truth joints over all valid future poses:
\begin{equation}
\mathrm{MPJPE}
=
\frac{1}{J_{\mathrm{pos}}N_{\mathrm{valid}}}
\sum_{n\in\mathcal D}
\sum_{k=1}^{K}
\sum_{j=1}^{J_{\mathrm{pos}}}
m_{n,k}
\left\|
\hat{\mathbf p}_{n,k,j}
-
\mathbf p_{n,k,j}
\right\|_2.
\end{equation}

PA-MPJPE evaluates
pose accuracy after removing global similarity misalignment. For each
valid pose of sample \(n\) at future step \(k\), we align the predicted joints to the ground truth using the optimal similarity transformation:
\begin{equation}
\begin{gathered}
\left(
s^\star_{n,k},
\mathbf Q^\star_{n,k},
\mathbf b^\star_{n,k}
\right)
=
\operatorname{ProcAlign}
\left(
\hat{\mathbf P}_{n,k},
\mathbf P_{n,k}
\right),\\[2pt]
\tilde{\mathbf P}_{n,k}
=
s^\star_{n,k}
\mathbf Q^\star_{n,k}
\hat{\mathbf P}_{n,k}
+
\mathbf b^\star_{n,k}
\mathbf 1^{\top},
\end{gathered}
\end{equation}
where
\(s^\star_{n,k}\in\mathbb{R}_{+}\),
\(\mathbf Q^\star_{n,k}\in \mathrm{SO}(3)\), and
\(\mathbf b^\star_{n,k}\in\mathbb{R}^{3}\) denote the optimal scale,
alignment rotation, and translation, respectively, and $\tilde{\mathbf P}_{n,k}$ denotes the aligned prediction.
The vector \(\mathbf 1\in\mathbb R^{J_{\mathrm{pos}}}\) broadcasts the translation to all joints.
Let \(\tilde{\mathbf p}_{n,k,j}\) denote the \(j\)-th column of
\(\tilde{\mathbf P}_{n,k}\). PA-MPJPE is then defined as:
\begin{equation}
\mathrm{PA\text{-}MPJPE}
=
\frac{
\sum_{n\in\mathcal D}
\sum_{k=1}^{K}
\sum_{j=1}^{J_{\mathrm{pos}}}
m_{n,k}
\left\|
\tilde{\mathbf p}_{n,k,j}
-
\mathbf p_{n,k,j}
\right\|_2
}{
J_{\mathrm{pos}}N_{\mathrm{valid}}
}.
\end{equation}

Root Trans. measures the Euclidean distance between predicted and ground-truth SMPL root translations, reflecting the accuracy of global body displacement.
Let \(\mathbf t_{n,k},\hat{\mathbf t}_{n,k}\in\mathbb{R}^{3}\) denote the ground-truth and predicted SMPL root translations for sample \(n\) at future step \(k\), respectively. 
It is defined as:
\begin{equation}
\text{Root Trans.}
=
\frac{1}{N_{\mathrm{valid}}}
\sum_{n\in\mathcal D}
\sum_{k=1}^{K}
m_{n,k}
\left\|
\hat{\mathbf t}_{n,k}
-
\mathbf t_{n,k}
\right\|_2.
\end{equation}

Body Geo. measures the rotational discrepancy of
the \(J_{\mathrm{rot}}=23\) non-root body joints. Let \(\mathbf R_{n,k,j},\hat{\mathbf R}_{n,k,j}\in \mathrm{SO}(3)\) denote the ground-truth and predicted local rotation matrices of body joint \(j\), respectively. Their geodesic angular distance is defined as:
\begin{equation}
d^{\mathrm{rot}}_{n,k,j}
=
\arccos\left(
\frac{
\operatorname{tr}
\left(
\hat{\mathbf R}_{n,k,j}^{\top}
\mathbf R_{n,k,j}
\right)-1
}{2}
\right),
\end{equation}
where 
the argument of
\(\arccos(\cdot)\) is clipped to \([-1,1]\) for numerical stability.
The mean body geodesic error, excluding the root joint, is
\begin{equation}
\text{Body Geo.}
=
\frac{180}{\pi J_{\mathrm{rot}}N_{\mathrm{valid}}}
\sum_{n\in\mathcal D}
\sum_{k=1}^{K}
\sum_{j=1}^{J_{\mathrm{rot}}}
m_{n,k}
d^{\mathrm{rot}}_{n,k,j}.
\end{equation}

To account for the inherent multimodality of future body motion, the model generates multiple plausible pose sequences for each observation. We therefore report Single and Best-5 to evaluate the first candidate and the best candidate among five forecasts, respectively.
For the \(q\)-th candidate, let \(\hat{\mathcal X}^{(q)}_n
=\{\hat{\mathbf X}^{(q)}_{n,k}\}_{k=1}^{K}\) denote the predicted SMPL sequence. Let \(\hat{\mathbf P}^{(q)}_{n,k}
\in\mathbb{R}^{3\times J_{\mathrm{pos}}}\) and
\(\hat{\mathbf p}^{(q)}_{n,k,j}\in\mathbb{R}^{3}\)
denote its joint coordinate matrix and \(j\)-th joint coordinate, respectively.  
For Best-5, we select for each sample the candidate with the lowest joint position error over all valid future steps:
\begin{equation}
q^\star_n
=
\arg\min_{q\in\{1,\ldots,5\}}
\sum_{k=1}^{K}
\sum_{j=1}^{J_{\mathrm{pos}}}
m_{n,k}
\left\|
\hat{\mathbf p}^{(q)}_{n,k,j}
-
\mathbf p_{n,k,j}
\right\|_2.
\end{equation}
All Best-5 pose metrics are computed from the same selected SMPL sequence \(\hat{\mathcal X}^{(q^\star_n)}_n\). Its joint coordinates 
are used for MPJPE and PA-MPJPE, while its root translations and body joint rotations are used for Root Trans. and Body Geo., respectively. This ensures that all Best-5 metrics evaluate a consistent motion forecast selected according to joint position accuracy.

\section{Method}

\begin{algorithm}[!t]
\caption{Training and inference of HIGFlow}
\label{alg:higflow}
\renewcommand{\algorithmicrequire}{\textbf{Input:}}
\renewcommand{\algorithmicensure}{\textbf{Output:}}
\footnotesize
\setlength{\algorithmicindent}{0.5em}
\begin{algorithmic}[1]

\REQUIRE Observed contexts
\((\mathcal C_{\mathrm{loc}},\mathcal H_{\mathrm{pose}})\),
ground-truth sequences
\((\mathcal Y,\mathcal X)\),
validity masks
\((\mathcal M_Y,\mathcal M_X)\),
forecasting horizon \(K\),
Heun steps \(N_{\mathrm{ODE}}\),
pose candidates \(S\).

\ENSURE Interaction location sequence
\(\hat{\mathcal Y}\)
and pose sequences
\(\{\hat{\mathcal X}^{(q)}\}_{q=1}^{S}\).

\item[] \textbf{// Training}

\item[] \textbf{Stage 1: Interaction location training}

\STATE Encode
\(\mathcal C_{\mathrm{loc}}\)
with Qwen3-VL and extract
\(\{\mathbf h_k^{\mathrm Q}\}_{k=1}^{K}\)
at the indexed
\mbox{\texttt{<hand\_traj\_k>}}
positions.

\STATE Decode the interaction locations and train the location predictor
by minimizing
\(\mathcal L_{\mathrm{loc}}\) in
Eq.~\eqref{eq:loc-loss}
using \(\mathcal M_Y\).

\STATE Encode the observed frames with V-JEPA.
Fuse the dynamic features with
\(\{\mathbf h_k^{\mathrm Q}\}_{k=1}^{K}\)
and update the V-JEPA adapter by minimizing
\(\lambda_{\mathrm{ADE}}\mathcal L_{\mathrm{ADE}}
+\lambda_{\mathrm{S1}}\mathcal L_{\mathrm{S1}}\)
using \(\mathcal M_Y\).

\item[] \textbf{Stage 2: Full-body pose training}

\STATE Encode
\((\mathcal H_{\mathrm{pose}},\mathcal Y)\)
and obtain the anchor sequence
\(\{\hat{\mathbf X}^{\mathrm a}_k\}_{k=1}^{K}\)
by Eq.~\eqref{eq:anchor-pred}.

\STATE Train the anchor predictor by minimizing
\(\mathcal L_{\mathrm{anchor}}\) in
Eq.~\eqref{eq:anchor-loss}
using \(\mathcal M_X\).

\STATE Compute the residual targets
\(\{\mathbf r_k^\ast\}_{k=1}^{K}\)
by Eq.~\eqref{eq:target-residual}
and construct
\(\{\mathbf c_k^{\mathrm{flow}},
\mathcal R_k\}_{k=1}^{K}\)
by Eqs.~\eqref{eq:flow-condition}
and \eqref{eq:residual-regulation}.

\FOR{\(k=1\) to \(K\)}
    \STATE Sample
    \(\tau\sim\mathcal U(0,1)\)
    and
    \(\mathbf z_k\sim
    \mathcal N(\mathbf 0,\sigma_r^2\mathbf I_{75})\).

    \STATE Compute the flow state
    \(\mathbf x_{\tau,k}\)
    as
    \((1-\tau)\mathbf z_k
    +\tau\mathcal B_{\boldsymbol{\rho}}(\mathbf r_k^\ast)\).

    \STATE Accumulate
    \(\mathcal L_{\mathrm{FM}}\)
    according to Eq.~\eqref{eq:flow-loss}
    and compute the direct and Heun rollout residuals.
\ENDFOR

\STATE Compose the residuals with the anchors and train the flow
model by minimizing
\(\mathcal L_{\mathrm{pose}}\) in
Eq.~\eqref{eq:pose-loss}
using \(\mathcal M_X\).

\item[] \textbf{// Inference}

\STATE Apply the trained location predictor to
\(\mathcal C_{\mathrm{loc}}\)
and obtain
\(\hat{\mathcal Y}
=
\{\hat{\mathbf y}_k\}_{k=1}^{K}\).

\STATE Condition on \((\mathcal H_{\mathrm{pose}},\hat{\mathcal Y})\) to obtain the anchor sequence and residual flow conditions
\(\{\hat{\mathbf X}^{\mathrm a}_k,
\mathbf c_k^{\mathrm{flow}},
\mathcal R_k\}_{k=1}^{K}\).

\FOR{\(q=1\) to \(S\)}
    \STATE Independently sample the residual priors
    \(\{\mathbf z_k^{(q)}\}_{k=1}^{K}\).

    \STATE Integrate the residual flows with
    \(N_{\mathrm{ODE}}\) Heun steps
    and compose them with the anchors
    to obtain
    \(\{\hat{\mathbf X}^{(q)}_k\}_{k=1}^{K}\).

    \STATE Collect the \(K\) predicted body states to form
    \(\hat{\mathcal X}^{(q)}
    =
    \{\hat{\mathbf X}^{(q)}_k\}_{k=1}^{K}\).
\ENDFOR

\STATE \textbf{return}
\(\hat{\mathcal Y}\)
and
\(\{\hat{\mathcal X}^{(q)}\}_{q=1}^{S}\).

\end{algorithmic}
\end{algorithm}

\subsection{Problem Formulation}

Given an observed egocentric context, we formulate continuous 4D interaction forecasting as a coupled \emph{where-to-how} prediction problem: first forecasting an ordered sequence of future interaction locations and then predicting the temporally aligned full-body motion conditioned on these locations.

For interaction location forecasting, the observed context is $\mathcal{C}_{\mathrm{loc}}
=
\left(
\mathcal{V}_{\mathrm{obs}},
\mathcal{E},
\mathcal{O}_{\mathrm{obs}},
\mathcal{T}
\right)$,
where \(\mathcal{V}_{\mathrm{obs}}\), \(\mathcal E\),
\(\mathcal O_{\mathrm{obs}}\), and \(\mathcal T\) denote the observed
egocentric frames, structured environment descriptors, observed
location history, and task prompt, respectively. The future
location sequence is predicted as
\begin{equation}
\hat{\mathcal Y}
=
f_{\mathrm{loc}}(\mathcal{C}_{\mathrm{loc}})
=
\{\hat{\mathbf y}_{k}\}_{k=1}^{K},
\label{eq:loc-pred}
\end{equation}
where 
\(K\) is the forecasting horizon and \(\hat{\mathbf y}_{k}\in\mathbb{R}^{3}\) denotes the predicted interaction location at future step \(k\) in the normalized coordinate representation. The corresponding ground-truth location sequence is denoted by \(\mathcal Y=\{\mathbf y_k\}_{k=1}^{K}\).

For full-body pose forecasting, the model takes the observed pose
history
\(\mathcal{H}_{\mathrm{pose}}\) together with a future interaction location
sequence \(\tilde{\mathcal Y}\):
\begin{equation}
\hat{\mathcal X}
=
f_{\mathrm{pose}}
\left(
\mathcal{H}_{\mathrm{pose}},
\tilde{\mathcal Y}
\right)
=
\{\hat{\mathbf X}_{k}\}_{k=1}^{K}.
\end{equation}
During training, \(\tilde{\mathcal Y}=\mathcal Y\) is the ground-truth location sequence, whereas during inference, \(\tilde{\mathcal Y}=\hat{\mathcal Y}\) is predicted by the first stage. The corresponding ground-truth pose sequence is denoted by \(\mathcal{X}=\{\mathbf{X}_k\}_{k=1}^{K}\), with the same validity mask as the interaction location targets, i.e., \(\mathcal{M}_X=\mathcal{M}_Y\). Each \(\hat{\mathbf X}_{k}\in\mathbb{R}^{147}\) represents an SMPL body state comprising a 6D root orientation, a 3D root translation, and 23 local body joint rotations represented in continuous 6D form. 

\subsection{HIGFlow Framework}
HIGFlow instantiates the above \emph{where-to-how} formulation with the
cascaded architecture shown in
Fig.~\ref{fig:framework}. It consists
of two specialized components: \textbf{Semantic-Dynamic Location
Forecasting} for predicting continuous future interaction locations,
and \textbf{Hand-Conditioned Residual Flow Matching} for generating
the corresponding full-body motion. The predicted location sequence
serves as an explicit geometric condition for motion forecasting,
establishing a direct dependency between location progression and
body motion realization. Algorithm~\ref{alg:higflow} summarizes the
training and inference procedures.

\subsubsection{\textbf{\textit{Semantic-Dynamic Location Forecasting}}}
The interaction location stage forecasts an ordered sequence of continuous 3D interaction locations by combining high-level semantic grounding with short-horizon visual dynamics. Specifically, Qwen3-VL~\cite{QwenTeam2025Qwen3VL} encodes the observed egocentric context and provides semantic representations for future steps, while a frozen V-JEPA~\cite{MurLabadia2026VJEPA21} encoder supplies complementary motion-sensitive features for dynamic refinement. 

\textbf{Semantic context encoding.}
Qwen3-VL receives the sampled egocentric frames, task prompt, observed location history, and structured environment descriptors. The location and environment encoders transform
\(\mathcal O_{\mathrm{obs}}\) and \(\mathcal E\) into dense features, which replace their corresponding placeholder embeddings before the
Qwen3-VL forward pass. For each future step \(k\), we extract the hidden state at the indexed placeholder \mbox{\texttt{<hand\_traj\_k>}} as the step-specific semantic representation \(\mathbf h_k^{\mathrm Q}\).


\textbf{Dynamic feature augmentation.}
Semantic representations provide task- and object-level grounding but may not sufficiently capture short-term hand-object dynamics that are important for precise spatial forecasting.
We therefore encode the observed video with 
V-JEPA~\cite{MurLabadia2026VJEPA21} and project its latent features into the Qwen hidden space, followed by resampling into \(M\) dynamic memory tokens. 
Each future step representation
\(\mathbf h_k^{\mathrm Q}\), augmented with a learned step embedding \(\mathbf s_k^{\mathrm{loc}}\), attends to this dynamic memory to obtain the motion context
\(\mathbf c_k\).
We then inject the dynamic information through a gated residual adapter:
\begin{equation}
\left\{
\begin{aligned}
\mathbf h_k^{\mathrm F}
&=
\mathbf h_k^{\mathrm Q}
+
g_k\Delta\mathbf h_k,\\
\Delta\mathbf h_k
&=
\alpha\tanh
\left(
D\!\left(
[\mathbf h_k^{\mathrm Q},
 \mathbf c_k,
 \mathbf s_k^{\mathrm{loc}}]
\right)
\right),\\
g_k
&=
\operatorname{sigmoid}
\left(
G\!\left(
[\mathbf h_k^{\mathrm Q},
 \mathbf c_k,
 \mathbf s_k^{\mathrm{loc}}]
\right)
\right)
\end{aligned}
\right.,
\label{eq:semantic-dynamic-fusion}
\end{equation}
where \(D\) and \(G\) denote lightweight residual and gating heads, respectively. The scalar
\(g_k\in(0,1)\) adaptively controls the contribution of the dynamic residual, while
\(\alpha>0\) bounds its magnitude.




\textbf{Continuous coordinate decoding.}
A coordinate decoder maps each fused representation
\(\mathbf h_k^{\mathrm F}\) directly to the normalized continuous 3D interaction
location \(\hat{\mathbf y}_k\). This explicit regression head separates continuous spatial prediction from the language generation interface, avoiding the need to represent metric coordinates as text tokens. Let \(B\) denote the mini-batch size and \(b\in\{1,\ldots,B\}\) index the samples. Let \((\mathcal M_Y)_{b,k}\in\{0,1\}\) indicate whether the interaction target \(\mathbf y_{b,k}\) is valid. We define the mask-aware ADE loss as:
\begin{equation}
\mathcal L_{\mathrm{ADE}}
=
\frac{1}{B}
\sum_{b=1}^{B}
\frac{
\sum_{k=1}^{K}
(\mathcal M_Y)_{b,k}
\left\|
\hat{\mathbf y}_{b,k}
-
\mathbf y_{b,k}
\right\|_2
}{
\sum_{k=1}^{K}
(\mathcal M_Y)_{b,k}
}.
\end{equation}
The Smooth L1 location loss is defined as:
\begin{equation}
\mathcal L_{\mathrm{S1}}
=
\frac{1}{B}
\sum_{b=1}^{B}
\frac{
\sum_{k=1}^{K}
(\mathcal M_Y)_{b,k}
\operatorname{SmoothL1}
\left(
\hat{\mathbf y}_{b,k},
\mathbf y_{b,k}
\right)
}{
\sum_{k=1}^{K}
(\mathcal M_Y)_{b,k}
}.
\end{equation}
The overall location objective is:
\begin{equation}
\mathcal L_{\mathrm{loc}}
=
\lambda_{\mathrm{ADE}}\mathcal L_{\mathrm{ADE}}
+
\lambda_{\mathrm{S1}}\mathcal L_{\mathrm{S1}}
+
\lambda_{\mathrm{CE}}\mathcal L_{\mathrm{CE}},
\label{eq:loc-loss}
\end{equation}
where \(\mathcal L_{\mathrm{CE}}\) denotes the auxiliary language modeling loss. The \(\operatorname{SmoothL1}\) loss is computed coordinate-wise and averaged over the three spatial dimensions. Both location regression losses are evaluated in the normalized coordinate space.

\subsubsection{\textbf{\textit{Hand-Conditioned Residual Flow Matching}}}
Given the observed pose history and ordered future interaction locations, the pose stage first forecasts a location-conditioned deterministic anchor and then models stochastic residuals around it using conditional flow matching.


\begin{table*}[!t]
\caption{Interaction location forecasting results on the Coherent4D dataset. HIGFlow and all baselines are evaluated under the same 3D setting, following the MMTwin 3D setting. All values are reported in millimeters, with lower values being better. Bold and underline denote the best and second-best results, respectively.}
\label{tab:hand_location_results}
\centering


\begin{tabular}{lccccccccc}
\toprule
Model
& \multicolumn{3}{>{\columncolor{bikepurple}}c}{\textbf{\emph{Health}}}
& \multicolumn{3}{>{\columncolor{cookingyellow}}c}{\textbf{\emph{Bike Repair}}}
& \multicolumn{3}{>{\columncolor{healthblue}}c}{\textbf{\emph{Cooking}}}\\
\cmidrule(lr){2-4}\cmidrule(lr){5-7}\cmidrule(lr){8-10}
& ADE$\downarrow$ & ADE$_{90}\downarrow$ & FDE$\downarrow$
& ADE$\downarrow$ & ADE$_{90}\downarrow$ & FDE$\downarrow$
& ADE$\downarrow$ & ADE$_{90}\downarrow$ & FDE$\downarrow$\\
\midrule

FIction
& \underline{40.30} & \underline{62.62} & \textbf{40.58}
& 88.55 & 143.21 & 98.61
& \underline{100.61} & \underline{184.63} & \underline{107.11}\\

Qwen3-VL
& 57.53 & 89.54 & 82.04
& 90.71 & 156.55 & 103.81
& 103.95 & 199.20 & 112.22\\

V-JEPA
& 45.18 & 69.49 & 46.26
& \underline{86.50} & \underline{142.43} & \underline{96.33}
& 102.33 & 194.80 & 107.35\\

Diff-IP3D
& 292.84 & 433.26 & 294.25
& 525.63 & 842.93 & 561.19
& 647.23 & 1268.28 & 662.71\\

MMTwin
& 290.45 & 437.88 & 285.21
& 429.71 & 690.80 & 480.29
& 513.75 & 957.17 & 556.40\\

\textbf{HIGFlow}
& \textbf{40.21} & \textbf{59.52} & \underline{41.44}
& \textbf{80.91} & \textbf{131.87} & \textbf{93.78}
& \textbf{93.46} & \textbf{178.20} & \textbf{104.26}\\


\bottomrule
\end{tabular}
\end{table*}

\textbf{Spatiotemporal conditioning and anchoring.}
To preserve the kinematic structure of the human body, each observed SMPL state is represented as a 24-node graph consisting of one root joint and 23 articulated body joints, with skeletal connections defining the graph edges. Graph propagation captures dependencies among physically connected body parts while preserving the SMPL topology.
The resulting node features are pooled into frame-level pose tokens and processed by a pose history Transformer, whose final classification token state \(\mathbf h\) summarizes the observed motion history. 

Given the interaction location sequence \(\tilde{\mathcal Y}\) defined in the
problem formulation, we set
\(\Delta\tilde{\mathbf y}_1=\mathbf 0\) and
\(\Delta\tilde{\mathbf y}_k
=
\tilde{\mathbf y}_k-\tilde{\mathbf y}_{k-1}\)
for \(k=2,\ldots,K\). The future condition encoder represents each
future step as:
\begin{equation}
\left\{
\begin{aligned}
&(\mathbf f_k)_{k=1}^{K}
=
\operatorname{Transformer}_{\mathrm{fut}}
\left(
(\boldsymbol{\eta}_k)_{k=1}^{K}
\right),\\[2pt]
&\boldsymbol{\eta}_k
=
\operatorname{MLP}_{\mathrm c}
\left(
[
\tilde{\mathbf y}_k,
\Delta\tilde{\mathbf y}_k,
k/K
]
\right)
\end{aligned}
\right.,
\end{equation}
where \(\boldsymbol{\eta}_k\) encodes the interaction location,
displacement, and relative future step, and
\(\mathbf f_k\) denotes the
corresponding context-aware representation.

For each future step, the deterministic anchor is predicted by combining
the global pose history representation with the corresponding future
location representation:
\begin{equation}
\hat{\mathbf X}_k^{\mathrm a}
=
\operatorname{MLP}_{\mathrm a}
\left(
[\mathbf h,\mathbf f_k]
\right).
\label{eq:anchor-pred}
\end{equation}
The anchor estimates the root translation, root orientation, and
articulated body configuration, providing a reference grounded in the interaction location for subsequent residual motion generation rather than serving as the final prediction.

The anchor predictor is optimized with
\begin{equation}
\begin{aligned}
\mathcal L_{\mathrm{anchor}}
={}&
\lambda_{\mathrm{6D}}\mathcal L_{\mathrm{6D}}
+
\lambda_{\mathrm{geo}}\mathcal L_{\mathrm{geo}}
+
\lambda_{\mathrm{rootgeo}}\mathcal L_{\mathrm{rootgeo}}
\\
&+
\lambda_{\mathrm{bodygeo}}\mathcal L_{\mathrm{bodygeo}}
+
\lambda_{\mathrm{trans}}\mathcal L_{\mathrm{trans}}
\\
&+
\lambda_{\mathrm{vel}}\mathcal L_{\mathrm{vel}}
+
\lambda_{\mathrm{joint}}\mathcal L_{\mathrm{joint}},
\end{aligned}
\label{eq:anchor-loss}
\end{equation}
where \(\mathcal L_{\mathrm{6D}}\) is the coordinate-wise
\(\ell_1\) loss over the 24 continuous 6D rotations.
\(\mathcal L_{\mathrm{geo}}\) measures the mean geodesic error over all
24 rotations, while \(\mathcal L_{\mathrm{rootgeo}}\) and
\(\mathcal L_{\mathrm{bodygeo}}\) separately supervise the root and
23 non-root body rotations. All framewise losses are evaluated only at
future steps marked valid by \(\mathcal M_X\).
Let \(\hat{\mathbf t}^{\mathrm a}_{b,k}\) and
\(\mathbf t_{b,k}\) denote the predicted and ground-truth root translations, and let
\(\hat{\mathbf p}^{\mathrm a}_{b,k,j}\) and
\(\mathbf p_{b,k,j}\) denote the corresponding 3D joint positions. 
The translation and joint position losses are
\begin{equation}
\begin{gathered}
\mathcal L_{\mathrm{trans}}
=
\frac{
\sum_{b=1}^{B}
\sum_{k=1}^{K}
(\mathcal M_X)_{b,k}
\left\|
\hat{\mathbf t}^{\mathrm a}_{b,k}
-
\mathbf t_{b,k}
\right\|_1
}{
3
\sum_{b=1}^{B}
\sum_{k=1}^{K}
(\mathcal M_X)_{b,k}
},
\\[2pt]
\mathcal L_{\mathrm{joint}}
=
\frac{
\sum_{b=1}^{B}
\sum_{k=1}^{K}
(\mathcal M_X)_{b,k}
\sum_{j=1}^{J_{\mathrm{pos}}}
\left\|
\hat{\mathbf p}^{\mathrm a}_{b,k,j}
-
\mathbf p_{b,k,j}
\right\|_2
}{
J_{\mathrm{pos}}
\sum_{b=1}^{B}
\sum_{k=1}^{K}
(\mathcal M_X)_{b,k}
}.
\end{gathered}
\end{equation}
The velocity loss \(\mathcal L_{\mathrm{vel}}\) further penalizes the
masked coordinate-wise \(\ell_1\) error between consecutive differences
of the predicted and ground-truth 147-dimensional SMPL states, using only valid consecutive-step pairs.


After anchor pretraining, the pose history encoder, future condition
encoder, and anchor predictor are frozen.

\textbf{Residual Flow Matching.}
To capture multiple feasible motions without deviating excessively from
the deterministic anchor, we model stochastic variations in an
anchor-relative residual space using conditional Flow
Matching~\cite{lipman2022flow}.
For each future step \(k\), the target
residual is defined as:
\begin{equation}
\mathbf r_k^\ast
=
\operatorname{Residual}
\left(
\hat{\mathbf X}^{\mathrm a}_k,
\mathbf X_k
\right),
\label{eq:target-residual}
\end{equation}
where \(\operatorname{Residual}\) computes rotational corrections for the root and body joints via the logarithmic map of relative rotations, and the root translation correction by subtraction. Concatenating the 3D root rotation correction, 3D root translation correction, and \(23\times3\) body rotation corrections yields a \(75\)-dimensional residual vector. Root translation residuals are computed and clipped in meters, and are added to the anchor root translations in meters during pose composition.

To regulate the magnitude of stochastic motion variations, we derive a
step-specific residual gate from the pose history, future location
condition, and deterministic anchor:
\begin{equation}
\begin{gathered}
\mathbf c_k^{\mathrm{flow}}
=
\left[
\mathbf h,\,
\mathbf f_k,\,
P_{\mathrm a}
\left(
\hat{\mathbf X}^{\mathrm a}_k
\right)
\right],\\[2pt]
\boldsymbol{\gamma}_k
=
\boldsymbol{\gamma}_{\max}
\odot
\operatorname{sigmoid}
\left(
\Gamma
\left(
\mathbf c_k^{\mathrm{flow}}
\right)
\right).
\end{gathered}
\label{eq:flow-condition}
\end{equation}
Here, \(P_{\mathrm a}\) projects the anchor state and
\(\Gamma\) predicts
three gate values corresponding to root rotation, root translation, and
body rotation. The vector
\(\boldsymbol{\gamma}_{\max}\in\mathbb R_+^3\) sets their maximum strengths. 
We further impose component-specific residual bounds. Let
\(\boldsymbol{\rho}
=
\operatorname{concat}
(\rho_{\mathrm R}\mathbf 1_3,
 \rho_{\mathrm t}\mathbf 1_3,
 \rho_{\mathrm B}\mathbf 1_{69})\),
where \(\rho_{\mathrm R}\), \(\rho_{\mathrm t}\), and
\(\rho_{\mathrm B}\) bound the root rotation, root translation, and
body rotation residuals, respectively.
Denoting element-wise clipping
to \([-\boldsymbol{\rho},\boldsymbol{\rho}]\) by \(\mathcal B_{\boldsymbol{\rho}}\), we define the step-specific
regulation operator as: 
\begin{equation}
\mathcal R_k(\mathbf x)
=
\mathcal B_{\boldsymbol{\rho}}
\left(
\operatorname{bcast}(\boldsymbol{\gamma}_k)
\odot
\mathbf x
\right),
\label{eq:residual-regulation}
\end{equation}
where \(\operatorname{bcast}\) expands the three gate values over
the corresponding \(3\), \(3\), and \(69\) residual dimensions. The
training target is bounded as 
\(\bar{\mathbf r}_k^\ast
=
\mathcal B_{\boldsymbol{\rho}}(\mathbf r_k^\ast)\),
while generated residuals are additionally modulated by \(\mathcal R_k\). This regulation constrains stochastic deviations from
the anchor while allowing their magnitude to adapt to each future
interaction step.

For
\(\tau\sim\mathcal U(0,1)\) and
\(\mathbf z_k\sim
\mathcal N(\mathbf 0,\sigma_r^2\mathbf I_{75})\), we construct the linear probability path:
\begin{equation}
\mathbf x_{\tau,k}
=
(1-\tau)\mathbf z_k
+
\tau\bar{\mathbf r}_k^\ast.
\end{equation}
The conditional velocity field is trained with
\begin{equation}
\mathcal L_{\mathrm{FM}}
=
\mathbb E
\left[
\left\|
v
\left(
\mathbf x_{\tau,k},
\tau,
\mathbf c_k^{\mathrm{flow}}
\right)
-
\left(
\bar{\mathbf r}_k^\ast-\mathbf z_k
\right)
\right\|_2^2
\right],
\label{eq:flow-loss}
\end{equation}
where \(v\) is the conditional residual velocity field. The expectation is taken over valid future steps, flow times, and
Gaussian prior samples.

For training-time endpoint supervision, we obtain a direct residual
estimate from an intermediate flow state as:
\begin{equation}
\tilde{\mathbf r}^{\mathrm{direct}}_k
=
\mathcal R_k
\left(
\mathbf x_{\tau,k}
+
(1-\tau)
v
\left(
\mathbf x_{\tau,k},
\tau,
\mathbf c_k^{\mathrm{flow}}
\right)
\right).
\end{equation}
We additionally perform a differentiable
\(N_{\mathrm{ODE}}\)-step Heun rollout from the Gaussian prior and
denote the regulated terminal residual by
\(\tilde{\mathbf r}^{\mathrm{roll}}_k\).

Finally, kinematic pose reconstruction composes the direct and rollout
residuals with the deterministic anchor through
\(\operatorname{Compose}\), which applies rotational corrections via the exponential map and adds the root translation correction. For
\(u\in\{\mathrm{direct},\mathrm{roll}\}\), we reconstruct
\begin{equation}
\hat{\mathbf X}_k^u
=
\operatorname{Compose}
\left(
\hat{\mathbf X}_k^{\mathrm a},
\tilde{\mathbf r}_k^u
\right),
\qquad
\hat{\mathcal X}^u
=
\{\hat{\mathbf X}_k^u\}_{k=1}^{K}.
\end{equation}
The corresponding endpoint pose loss is:
\begin{equation}
\mathcal L_u
=
\mathcal E_{\mathrm{pose}}
\left(
\hat{\mathcal X}^u,
\mathcal X
\right),
\qquad
u\in\{\mathrm{direct},\mathrm{roll}\},
\end{equation}
where \(\mathcal E_{\mathrm{pose}}\) uses the same masked 6D rotation, geodesic rotation, root translation, temporal velocity, and 3D joint position terms as the anchor objective.

We further impose a residual alignment loss that encourages the direct endpoint residual to match the clipped target residual:
\begin{equation}
\mathcal L_{\mathrm{res}}
=
\frac{
\sum_b\sum_{k=1}^{K}
(\mathcal M_X)_{b,k}
\left\|
\tilde{\mathbf r}^{\mathrm{direct}}_{b,k}
-
\bar{\mathbf r}_{b,k}^\ast
\right\|_2^2
}{
75
\sum_b\sum_{k=1}^{K}
(\mathcal M_X)_{b,k}
}.
\end{equation}

The overall pose objective is:
\begin{equation}
\mathcal L_{\mathrm{pose}}
=
\lambda_{\mathrm{FM}}\mathcal L_{\mathrm{FM}}
+
\lambda_{\mathrm{direct}}\mathcal L_{\mathrm{direct}}
+
\lambda_{\mathrm{roll}}\mathcal L_{\mathrm{roll}}
+
\lambda_{\mathrm{res}}\mathcal L_{\mathrm{res}}.
\label{eq:pose-loss}
\end{equation}

At inference, we independently sample \(S\) residual priors and integrate each using 
\(N_{\mathrm{ODE}}\) Heun steps. The resulting regulated terminal residuals are composed with the deterministic anchor to produce \(S\) plausible future motion sequences,
\(\{\hat{\mathcal X}^{(q)}\}_{q=1}^{S}\).

\section{Experiments}
\subsection{Experimental Settings}

\begin{table*}[!t]
\caption{
Full-body pose forecasting results on the Coherent4D dataset.
The left side reports pose forecasting conditioned on ground-truth future interaction locations, while the right side reports pose forecasting conditioned on predicted future interaction locations.
MPJPE, PA-MPJPE, and Root Trans. are reported in millimeters, while Body Geo. is reported in degrees.
Single uses the first candidate, whereas Best-5 reports all metrics for the candidate selected by sample-level MPJPE.
Lower values are better.
Bold and underline denote the best and second-best results, respectively.
}
\label{tab:pose_forecasting_all_combined}
\centering
\setlength{\tabcolsep}{2.8pt}

\begin{tabular}{lcccccccc|cccccccc}
\toprule

Model
& \multicolumn{2}{c}{MPJPE$\downarrow$}
& \multicolumn{2}{c}{PA-MPJPE$\downarrow$}
& \multicolumn{2}{c}{Root Trans.$\downarrow$}
& \multicolumn{2}{c|}{Body Geo.$\downarrow$}
& \multicolumn{2}{c}{MPJPE$\downarrow$}
& \multicolumn{2}{c}{PA-MPJPE$\downarrow$}
& \multicolumn{2}{c}{Root Trans.$\downarrow$}
& \multicolumn{2}{c}{Body Geo.$\downarrow$} \\

\cmidrule(lr){2-3}
\cmidrule(lr){4-5}
\cmidrule(lr){6-7}
\cmidrule(lr){8-9}
\cmidrule(lr){10-11}
\cmidrule(lr){12-13}
\cmidrule(lr){14-15}
\cmidrule(lr){16-17}

& Single & Best-5
& Single & Best-5
& Single & Best-5
& Single & Best-5
& Single & Best-5
& Single & Best-5
& Single & Best-5
& Single & Best-5 \\

\midrule

\rowcolor{bikepurple}
\multicolumn{17}{@{}c@{}}{\textbf{\emph{Health}}}\\

FIction
& \underline{117.17} & 115.91
& \textbf{42.12} & \underline{42.01}
& \underline{104.09} & 102.91
& 9.03 & 9.02
& \underline{124.59} & 123.43
& 50.82 & 49.73
& \underline{108.67} & 107.52
& \underline{9.01} & \underline{9.00}
\\

SkeletonDiffusion
& 166.32 & 137.39
& 53.18 & 52.19
& 158.85 & 138.47
& 10.55 & 10.45
& 184.97 & 155.55
& 55.01 & 54.20
& 172.37 & 151.19
& 10.89 & 10.81
\\

SLD-HMP
& 421.56 & \textbf{88.38}
& 44.79 & \textbf{32.50}
& 403.30 & \textbf{74.40}
& \textbf{8.76} & \textbf{5.63}
& 402.76 & \textbf{92.15}
& \underline{48.66} & \textbf{33.22}
& 381.66 & \textbf{89.37}
& \textbf{8.56} & \textbf{5.72}
\\

\textbf{HIGFlow}
& \textbf{95.25} & \underline{93.45}
& \underline{43.79} & 43.66
& \textbf{80.93} & \underline{79.06}
& \underline{8.80} & \underline{8.79}
& \textbf{119.96} & \underline{118.14}
& \textbf{47.06} & \underline{46.99}
& \textbf{94.19} & \underline{92.29}
& 9.33 & 9.32
\\

\midrule

\rowcolor{cookingyellow}
\multicolumn{17}{@{}c@{}}{\textbf{\emph{Bike Repair}}}\\
\addlinespace[0.2em]

FIction
& 295.34 & 284.52
& 78.95 & \underline{78.24}
& 292.07 & 279.21
& \underline{13.60} & \underline{13.50}
& \underline{351.64} & 341.06
& 86.75 & 85.85
& 325.06 & 312.04
& \underline{14.14} & \underline{14.03}
\\

SkeletonDiffusion
& 453.52 & 370.26
& 113.22 & 109.63
& 405.49 & 331.24
& 17.69 & 17.44
& 469.82 & 398.82
& 115.09 & 111.77
& 408.17 & 343.07
& 17.89 & 17.74
\\

SLD-HMP
& \underline{269.06} & \underline{234.95}
& \underline{74.34} & \textbf{70.03}
& \underline{261.00} & \textbf{218.63}
& \textbf{11.97} & \textbf{11.12}
& 358.20 & \underline{314.88}
& \underline{83.82} & \textbf{81.76}
& \underline{311.54} & \underline{296.01}
& \textbf{12.39} & \textbf{11.76}
\\

\textbf{HIGFlow}
& \textbf{213.65} & \textbf{210.09}
& \textbf{70.07} & \textbf{70.03}
& \textbf{224.19} & \underline{220.18}
& 15.05 & 15.05
& \textbf{306.92} & \textbf{303.07}
& \textbf{82.26} & \underline{82.25}
& \textbf{287.24} & \textbf{282.98}
& 16.07 & 16.07
\\

\midrule

\rowcolor{healthblue}
\multicolumn{17}{@{}c@{}}{\textbf{\emph{Cooking}}}\\
\addlinespace[0.2em]

FIction
& 237.86 & 226.62
& 43.81 & 43.76
& 227.23 & 214.75
& 8.34 & 8.34
& 374.89 & 364.15
& 46.30 & 46.28
& 348.70 & 336.89
& 8.51 & 8.51
\\

SkeletonDiffusion
& 310.39 & 260.00
& 51.43 & 51.25
& 278.05 & 229.97
& 11.42 & 11.41
& 421.75 & 382.35
& 53.09 & 52.91
& 377.64 & 339.03
& 11.89 & 11.88
\\

SLD-HMP
& \underline{184.42} & \underline{167.19}
& \underline{43.41} & \underline{41.21}
& \underline{174.33} & \underline{156.25}
& \underline{7.67} & \textbf{7.25}
& \underline{358.92} & \underline{342.11}
& \textbf{44.86} & \textbf{43.13}
& \underline{332.74} & \underline{313.26}
& \textbf{7.83} & \textbf{7.53}
\\

\textbf{HIGFlow}
& \textbf{143.98} & \textbf{143.41}
& \textbf{39.45} & \textbf{39.42}
& \textbf{139.26} & \textbf{138.73}
& \textbf{7.53} & \underline{7.53}
& \textbf{340.72} & \textbf{340.24}
& \underline{45.65} & \underline{45.64}
& \textbf{312.56} & \textbf{312.11}
& \underline{8.50} & \underline{8.50}
\\

\bottomrule
\end{tabular}
\end{table*}

\subsubsection{\textbf{\textit{Baselines}}}
We compare HIGFlow with two groups of baselines corresponding to the two forecasting stages. 

For interaction location forecasting, we consider FIction~\cite{ashutosh2025fiction}, Qwen3-VL-2B~\cite{QwenTeam2025Qwen3VL}, V-JEPA 2.1~\cite{MurLabadia2026VJEPA21}, Diff-IP3D~\cite{ma2025diff}, and MMTwin~\cite{ma2025novel}. FIction is the closest prior method coupling interaction localization with pose forecasting, while Qwen3-VL and V-JEPA 2.1 provide semantic- and dynamics-oriented baselines, respectively. Following MMTwin~\cite{ma2025novel}, which extends Diff-IP2D~\cite{ma2025diff} and MADiff~\cite{ma2026madiff} from 2D to 3D, we extend Diff-IP2D to predict continuous 3D interaction locations and refer to this variant as Diff-IP3D. MMTwin is reproduced in its 3D configuration with its prediction targets matched to Coherent4D. All methods are evaluated in the shared sample-local coordinate frame using the corrected USST evaluation implementation~\cite{bao2023uncertainty}.\footnote{See the USST erratum commit: \url{https://github.com/oppo-us-research/USST/commit/beebdb963a702b08de3a4cf8d1ac9924b544abc4}.}

For full-body pose forecasting, we compare HIGFlow with FIction, SkeletonDiffusion, and SLD-HMP under the same interaction location conditioning protocol. During training, all methods receive ground-truth future interaction locations. FIction follows its original location-conditioned formulation, whereas SkeletonDiffusion and SLD-HMP retain their original forecasting architectures with only the input and output interfaces adapted to our SMPL representation. We report both Single and Best-5 results following the evaluation protocol defined above.

\subsubsection{\textbf{\textit{Implementation Details}}}

Throughout this paper, V-JEPA refers to V-JEPA 2.1 ViT-Giant/384, and
Qwen3-VL to Qwen3-VL-2B. For interaction location forecasting,
we use \(M=64\) resampled V-JEPA memory
tokens, set \(\alpha=0.1\), and employ a three-layer coordinate decoder. The loss weights are
\((\lambda_{\mathrm{ADE}},\lambda_{\mathrm{S1}},
\lambda_{\mathrm{CE}})=(1.5,0.6,0.01)\).
The base predictor and JEPA adapter are trained sequentially with
AdamW and cosine learning rate schedules, using learning rate/warmup ratio pairs of \((1\times10^{-5},0.10)\) and
\((3\times10^{-6},0.05)\), respectively, while the base predictor,
structured encoders, and coordinate decoder remain frozen during
adapter training.

The pose history and future condition Transformers contain four and
two layers, respectively, while \(\operatorname{MLP}_{\mathrm c}\) and
\(\operatorname{MLP}_{\mathrm a}\) use two and three linear layers.
The anchor predictor is pretrained with AdamW for up to 60 epochs, using
a learning rate of \(8\times10^{-5}\), weight decay \(10^{-4}\),
gradient clipping at \(1.0\), and an early stopping patience of six
epochs. 
For Health and Cooking, we set
\((\lambda_{\mathrm{6D}},\lambda_{\mathrm{geo}},
\lambda_{\mathrm{rootgeo}},\lambda_{\mathrm{bodygeo}},
\lambda_{\mathrm{vel}},\lambda_{\mathrm{joint}})
=
(1,0.1,0.5,1,1,10)\),
with \(\lambda_{\mathrm{trans}}=12\), batch size \(1\), and dropout
\(0.10\).
For Bike Repair, we set
\(\lambda_{\mathrm{rootgeo}}=\lambda_{\mathrm{bodygeo}}=0\) and \(\lambda_{\mathrm{trans}}=10\), with the remaining loss weights unchanged, a batch size of \(12\), and dropout \(0.15\).

Residual flow matching freezes the pose history encoder,
future condition encoder, and anchor predictor. We use
\(\sigma_r=0.003\), batch size \(16\), and AdamW with weight decay
\(10^{-4}\). The learning rates are \(8\times10^{-5}\) for Health and
Cooking and \(5\times10^{-5}\) for Bike Repair. The loss weights are set to
\(\lambda_{\mathrm{FM}}=\lambda_{\mathrm{direct}}
=\lambda_{\mathrm{roll}}=1.0\) and
\(\lambda_{\mathrm{res}}=0.5\).
For Health and Cooking, we set
\(\boldsymbol{\gamma}_{\max}=(1.0,1.0,0.7)\) and
\((\rho_{\mathrm R},\rho_{\mathrm t},\rho_{\mathrm B})
=(0.174533\,\mathrm{rad},0.12\,\mathrm m,0.12\,\mathrm{rad})\).
For Bike Repair,
the corresponding settings are \(\boldsymbol{\gamma}_{\max}=(0.8,1.0,0.35)\) and
\((\rho_{\mathrm R},\rho_{\mathrm t},\rho_{\mathrm B})
=(0.25\,\mathrm{rad},0.25\,\mathrm m,0.08\,\mathrm{rad})\).
The bound \(\rho_t\) is applied directly to each coordinate of the root translation residual in meters, without division by \(s\). 
Both training and inference use \(N_{\mathrm{ODE}}=4\) Heun steps. At inference, we generate \(S=5\)
candidates for Best-5 evaluation. All experiments are conducted on
six NVIDIA A100 GPUs.

\begin{table}[!t]
\caption{Ablation of V-JEPA residual fusion for interaction location forecasting. LocEnc, EnvEnc, CoordDec, and sampled video frames are fixed. All values are reported in millimeters, and lower values are better.}
\label{tab:jepa_injection_ablation}
\centering
\footnotesize
\providecommand{\cmark}{\ensuremath{\surd}}
\setlength{\tabcolsep}{3.0pt}
\renewcommand{\arraystretch}{1.10}
\begin{tabular}{@{}lc|ccc@{}}
\toprule
Domain
& V-JEPA injection
& ADE$\downarrow$
& ADE$_{90}\downarrow$
& FDE$\downarrow$ \\
\midrule

\multirow{2}{*}{Health}
& --
& 40.28 & \textbf{59.34} & 41.47 \\
& \cmark
& \textbf{40.21} & 59.52 & \textbf{41.44} \\

\midrule
\multirow{2}{*}{Bike Repair}
& --
& 81.18 & 134.10 & 94.03 \\
& \cmark
& \textbf{80.91} & \textbf{131.87} & \textbf{93.78} \\

\midrule
\multirow{2}{*}{Cooking}
& --
& 94.20 & 178.21 & 104.83 \\
& \cmark
& \textbf{93.46} & \textbf{178.20} & \textbf{104.26} \\

\bottomrule
\end{tabular}
\end{table}

\subsection{Quantitative Analysis}




We organized the quantitative analysis around the following questions.

\textbf{\textit{1) Can HIGFlow jointly forecast interaction locations and full-body poses while outperforming specialized baselines?}}

We evaluated the two stages of HIGFlow against methods designed specifically for the corresponding forecasting tasks. Table~\ref{tab:hand_location_results} reports interaction location forecasting results, while Table~\ref{tab:pose_forecasting_all_combined} reports full-body pose forecasting results under two location-conditioning settings: the left part uses ground-truth future interaction locations, whereas the right part uses predicted future interaction locations and represents the standard forecasting setting. For each task and conditioning setting, HIGFlow and all corresponding baselines are evaluated under the same evaluation protocol, enabling direct comparison of their forecasting performance. HIGFlow achieves leading performance in both interaction location forecasting and full-body pose forecasting, with particularly strong results on Cooking and Bike Repair and competitive performance on Health. Overall, HIGFlow demonstrates consistent performance on the interaction forecasting task.

\begin{table}[!t]
\caption{Domain-wise motion statistics in the shared sample-local
coordinate frame. Loc. Disp. and Root Disp. measure the mean displacement
between consecutive valid future steps for interaction locations and SMPL
root translations, respectively. Values are reported in millimeters.}
\label{tab:domain_motion_range}
\setlength{\tabcolsep}{4.0pt}
\begin{tabular}{lccc}
\toprule
Domain & Loc. Disp. & Root Disp. & Spatial pattern\\
\midrule
Health & 153.86 & 59.45 & Compact spatial movement\\
Bike Repair & 293.53 & 195.71 & Tool-centric reaching and repair\\
Cooking & 318.69 & 171.48 & Broad object-centric movement\\
\bottomrule
\end{tabular}
\end{table}

\begin{table}[!t]
\caption{Effect of the V-JEPA memory token budget on interaction location
forecasting. \(M\) denotes the number of resampled V-JEPA memory tokens.
The shaded entries indicate the token budget selected for the final model.
All values are reported in millimeters, and lower values are better.
Bold denotes the best result for each domain and metric.}
\label{tab:jepa_token_budget}
\centering
\begin{tabular}{lcccc}
\toprule
Domain
& \(M\)
& ADE$\downarrow$
& ADE$_{90}\downarrow$
& FDE$\downarrow$ \\
\midrule

\multirow{4}{*}{Health}
& 16
& 40.23
& 59.61
& 41.46 \\
& 32
& 40.23
& 59.49
& 41.46 \\
& \cellcolor{jepagrey}64
& \cellcolor{jepagrey}\textbf{40.21}
& \cellcolor{jepagrey}59.52
& \cellcolor{jepagrey}41.44 \\
& 128
& 40.23
& \textbf{59.44}
& \textbf{41.42} \\

\midrule

\multirow{4}{*}{Bike Repair}
& 16
& 88.47
& 138.75
& 98.59 \\
& 32
& 87.08
& 136.32
& 97.56 \\
& \cellcolor{jepagrey}64
& \cellcolor{jepagrey}\textbf{80.91}
& \cellcolor{jepagrey}\textbf{131.87}
& \cellcolor{jepagrey}\textbf{93.78} \\
& 128
& 86.94
& 135.53
& 97.58 \\

\midrule

\multirow{4}{*}{Cooking}
& 16
& 93.36
& 178.45
& \textbf{104.20} \\
& 32
& \textbf{93.04}
& \textbf{177.63}
& 104.33 \\
& \cellcolor{jepagrey}64
& \cellcolor{jepagrey}93.46
& \cellcolor{jepagrey}178.20
& \cellcolor{jepagrey}104.26 \\
& 128
& 93.36
& 178.03
& 104.22 \\

\bottomrule
\end{tabular}
\end{table}

\begin{table*}[!t]
\caption{Ablation of input representations and output decoding for interaction location forecasting. LocEnc, EnvEnc, and CoordDec denote the location encoder, environment encoder, and coordinate decoder, respectively. A dash in the LocEnc or EnvEnc column indicates that the corresponding input is represented as text rather than processed by a dedicated encoder. A dash in the CoordDec column indicates that interaction locations are generated as text rather than decoded by a dedicated coordinate decoder. Gray rows include V-JEPA residual fusion. All values are reported in millimeters, with lower values indicating better performance and the best results highlighted in bold.}
\label{tab:hand_format_ablation}
\centering
\providecommand{\cmark}{\ensuremath{\surd}}
\setlength{\tabcolsep}{5pt}
\begin{tabular}{lccc|ccc|ccc|ccc}
\toprule
Variant & LocEnc & EnvEnc & CoordDec
& \multicolumn{3}{c|}{Health}
& \multicolumn{3}{c|}{Bike Repair}
& \multicolumn{3}{c}{Cooking} \\
\cmidrule(lr){5-7}
\cmidrule(lr){8-10}
\cmidrule(lr){11-13}
& & & 
& ADE$\downarrow$ & ADE$_{90}\downarrow$ & FDE$\downarrow$
& ADE$\downarrow$ & ADE$_{90}\downarrow$ & FDE$\downarrow$
& ADE$\downarrow$ & ADE$_{90}\downarrow$ & FDE$\downarrow$ \\
\midrule

Qwen3-VL
& -- & -- & --
& 57.53 & 89.54 & 82.04
& 90.71 & 156.55 & 103.81
& 103.95 & 199.20 & 112.22 \\

w/o CoordDec 
& \cmark & \cmark & --
& 57.11 & 89.31 & 59.71
& 99.64 & 161.61 & 113.42
& 113.87 & 222.51 & 119.68 \\

w/o LocEnc 
& -- & \cmark & \cmark
& 42.57 & 63.41 & 44.88
& 87.78 & 143.79 & 98.04
& 97.29 & 182.26 & 105.15 \\

\rowcolor{jepagrey}
w/o LocEnc + V-JEPA 
& -- & \cmark & \cmark
& 42.18 & 63.11 & 44.20
& 86.40 & 139.00 & 96.76
& \textbf{92.14} & 179.32 & \textbf{101.24} \\

w/o EnvEnc 
& \cmark & -- & \cmark
& 42.17 & 65.26 & 43.32
& 87.65 & 137.12 & 97.31
& 100.13 & 184.76 & 106.65 \\

\rowcolor{jepagrey}
w/o EnvEnc + V-JEPA 
& \cmark & -- & \cmark
& 41.86 & 63.68 & 42.83
& 85.94 & 133.97 & 95.19
& 96.24 & 182.77 & 103.39 \\

\midrule

\multicolumn{4}{@{}l|}{\textbf{HIGFlow (Full model)}}
& \textbf{40.21} & \textbf{59.52} & \textbf{41.44}
& \textbf{80.91} & \textbf{131.87} & \textbf{93.78}
& 93.46 & \textbf{178.20} & 104.26 \\

\bottomrule
\end{tabular}
\end{table*}

\textbf{\textit{2) What is the effect of V-JEPA residual fusion?}}

We fixed the location encoder, environment encoder, coordinate decoder, frame input, and training configuration, and varied only the JEPA-guided residual adapter. When enabled, the resampled V-JEPA memory tokens are fused with the hidden states at the future step readout positions after the Qwen3-VL decoder forward pass.
Table~\ref{tab:jepa_injection_ablation} shows that V-JEPA residual fusion generally improves interaction location forecasting, with the most pronounced gains on Bike Repair. These results indicate that V-JEPA provides complementary motion-sensitive visual cues that help localize future locations beyond the semantic and contextual information captured by the base predictor.

\textbf{\textit{3) Why do HIGFlow's gains vary across domains?}}

HIGFlow achieves larger improvements on Cooking and Bike Repair than on Health in both interaction location and pose forecasting. As shown in Table~\ref{tab:domain_motion_range}, Cooking and Bike Repair exhibit substantially greater location and root displacements, providing richer geometric and dynamic cues for modeling future locations and full-body motion. In contrast, Health involves more compact spatial movements, for which the competing methods already perform relatively well, leaving less room for improvement. These observations suggest that HIGFlow benefits particularly from domains involving larger and more complex motion variations.

\begin{table*}[!t]
\caption{Input ablation for interaction location forecasting across the three domains. Loc. and Env. denote location history and environment context, respectively. Frames denotes sampled video frames provided to Qwen3-VL. Gray rows include V-JEPA residual fusion, which retains features extracted by V-JEPA from the observed video. All values are reported in millimeters, with lower values indicating better performance and the best results highlighted in bold.}
\label{tab:hand_modality_ablation}
\providecommand{\cmark}{\ensuremath{\surd}}
\begin{tabular}{lccc|ccc|ccc|ccc}
\toprule
\multirow{2}{*}{Input}
& \multirow{2}{*}{Loc.}
& \multirow{2}{*}{Env.}
& \multirow{2}{*}{Frames}
& \multicolumn{3}{c|}{Health}
& \multicolumn{3}{c|}{Bike Repair}
& \multicolumn{3}{c}{Cooking} \\
\cmidrule(lr){5-7}
\cmidrule(lr){8-10}
\cmidrule(lr){11-13}
& & &
& ADE$\downarrow$ & ADE$_{90}\downarrow$ & FDE$\downarrow$
& ADE$\downarrow$ & ADE$_{90}\downarrow$ & FDE$\downarrow$
& ADE$\downarrow$ & ADE$_{90}\downarrow$ & FDE$\downarrow$ \\
\midrule

w/o Env.
& \cmark & -- & \cmark
& 37.40 & \textbf{58.16} & \textbf{38.65}
& 114.51 & 161.50 & 124.01
& 97.92 & 189.73 & 106.63 \\

\rowcolor{jepagrey}
w/o Env. + V-JEPA
& \cmark & -- & \cmark
& \textbf{37.32} & 58.27 & 38.68
& 91.79 & 141.30 & 101.61
& 94.60 & 186.08 & 104.71 \\

w/o Loc.
& -- & \cmark & \cmark
& 69.48 & 103.51 & 67.04
& 135.71 & 199.00 & 144.62
& 143.82 & 267.09 & 148.47 \\

\rowcolor{jepagrey}
w/o Loc. + V-JEPA
& -- & \cmark & \cmark
& 55.59 & 85.30 & 55.53
& 162.59 & 250.96 & 168.90
& 127.97 & 246.60 & 130.92 \\

w/o Frames
& \cmark & \cmark & --
& 51.91 & 75.49 & 51.98
& 143.90 & 224.01 & 143.32
& 145.86 & 221.17 & 143.07 \\

\rowcolor{jepagrey}
w/o Frames + V-JEPA
& \cmark & \cmark & --
& 50.61 & 75.93 & 51.78
& 117.27 & 175.46 & 125.37
& 119.60 & 225.82 & 123.65 \\

\midrule

\multicolumn{4}{l|}{\textbf{HIGFlow (Full model)}}
& 40.21 & 59.52 & 41.44
& \textbf{80.91} & \textbf{131.87} & \textbf{93.78}
& \textbf{93.46} & \textbf{178.20} & \textbf{104.26} \\

\bottomrule
\end{tabular}
\end{table*}

\textbf{\textit{4) How do location stage design choices affect interaction location forecasting?}}

We examined three design aspects of the location stage: structured representation and coordinate decoding, input composition, and V-JEPA memory capacity. The results collectively support the use of structured multimodal cues, explicit continuous coordinate regression, and a moderate dynamic memory budget.

\textbf{a) Representation and coordinate decoding.}
We ablated the location encoder, environment encoder, and coordinate decoder while keeping the remaining components fixed. Without LocEnc or EnvEnc, the corresponding structured inputs are serialized as text. Without CoordDec, interaction locations are generated through the language interface rather than continuous coordinate regression. We also included Qwen3-VL as a baseline without these three components. As shown in Table~\ref{tab:hand_format_ablation}, CoordDec provides the clearest and most consistent gains, highlighting the benefit of directly regressing continuous 3D coordinates. LocEnc and EnvEnc generally improve cross-domain consistency by preserving structured location and scene information, although Cooking shows mixed sensitivity. Overall, the full model achieves the most balanced performance across domains.



\begin{figure*}[!t]
\centering

\includegraphics[width=0.95\textwidth]
{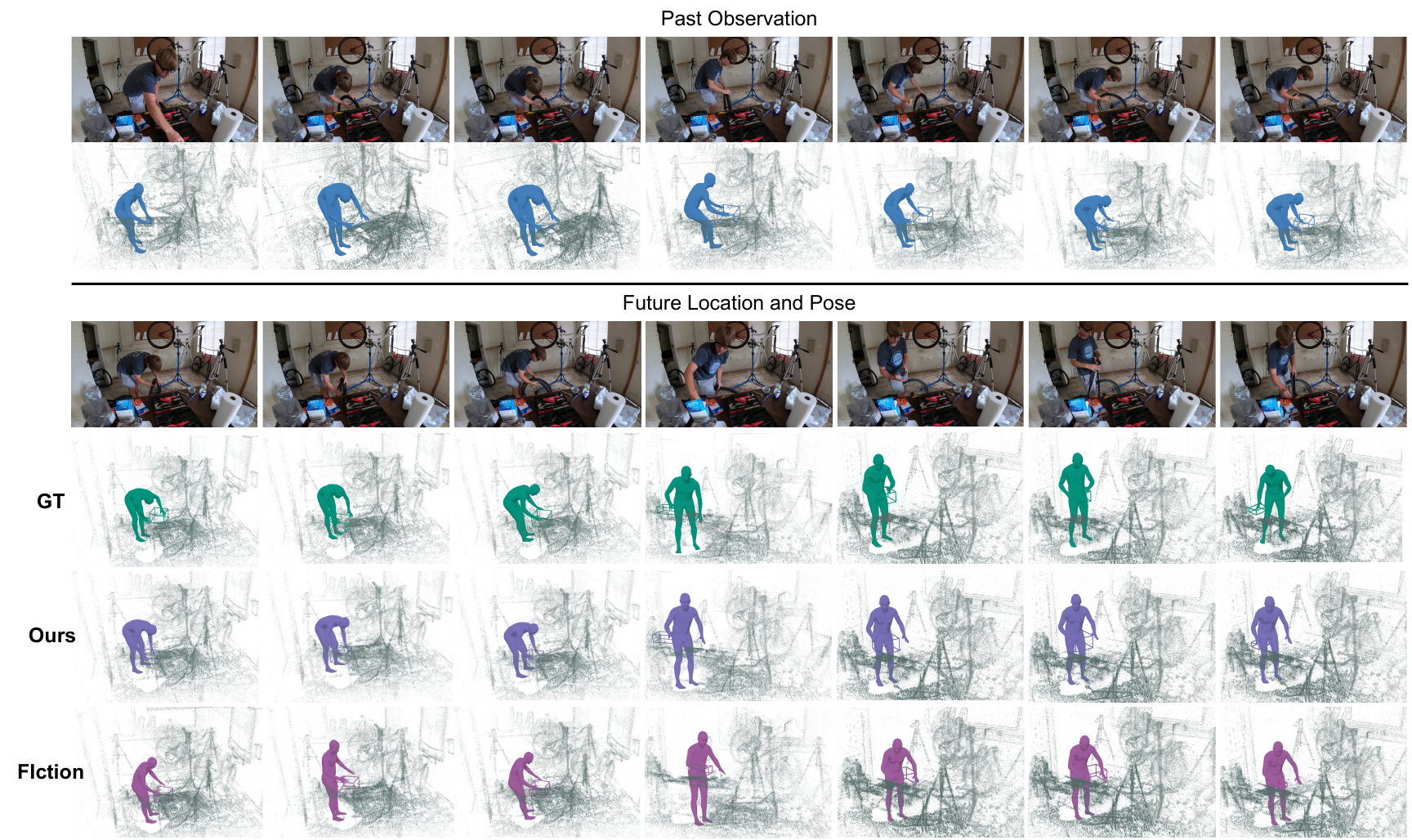}
\par\vspace{2mm}
{\footnotesize\bfseries (a) Bike Repair\par}

\vspace{4mm}

\includegraphics[width=0.95\textwidth]
{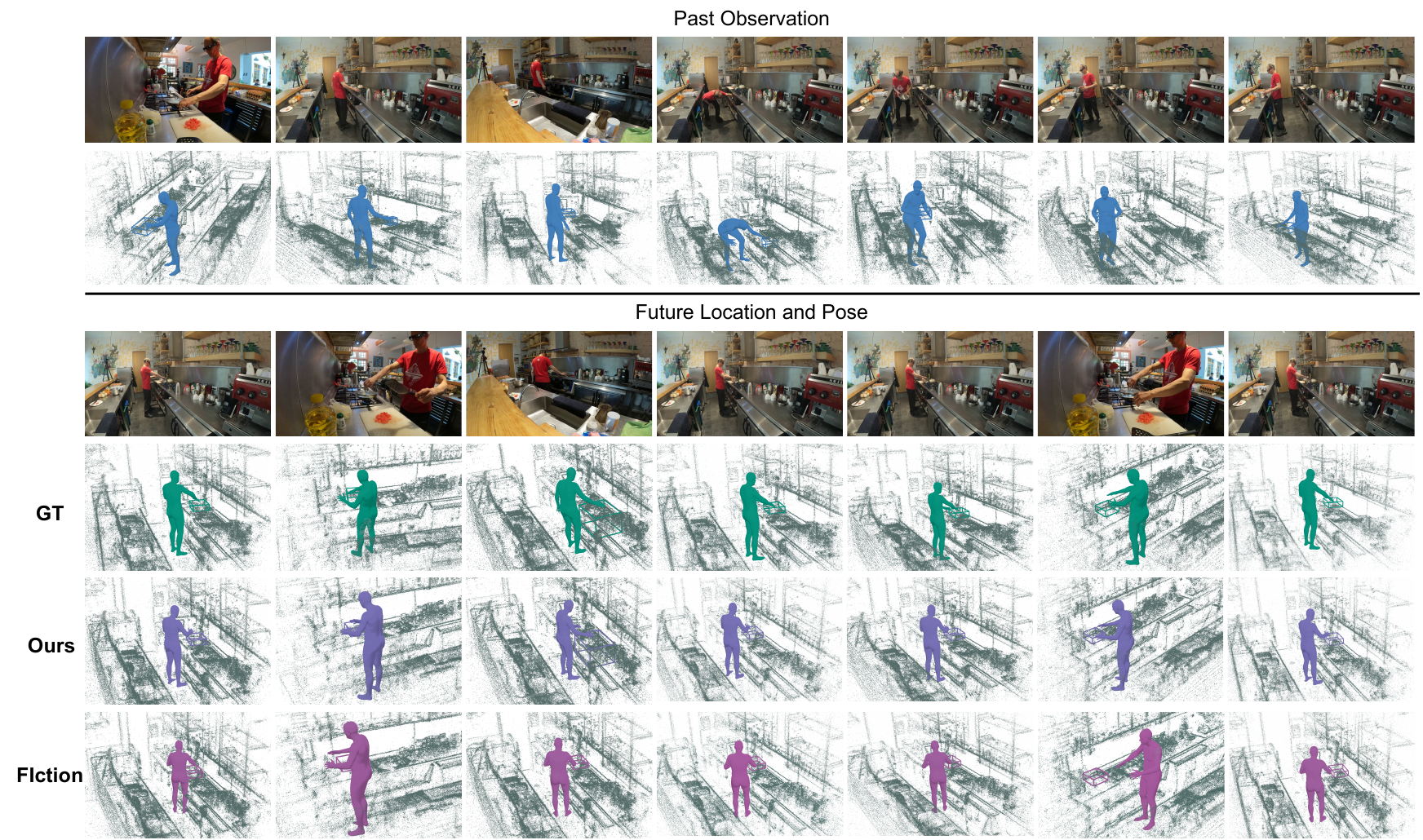}
\par\vspace{2mm}
{\footnotesize\bfseries (b) Cooking\par}

\caption{Qualitative pose forecasting comparison on Bike Repair and Cooking
samples. For each example, the upper panel shows the observed SMPL pose
history with synchronized exocentric source frames, while the lower panel
presents future exocentric reference frames together with the corresponding
ground-truth, HIGFlow, and FIction pose sequences. The exocentric frames are
included only for visualization and are not used as inputs to the pose models.}
\label{fig:Visualization}
\end{figure*}

\textbf{b) Input composition.}
We separately removed location history, environment context, or the sampled video frames provided to Qwen3-VL, while retaining the remaining inputs and coordinate decoding architecture. Each configuration was evaluated with and without V-JEPA residual fusion, which still provided features extracted from the observed video when Qwen3-VL received no frames. As shown in Table~\ref{tab:hand_modality_ablation}, removing location history or frame input substantially degrades forecasting performance, while environment context is particularly important for Bike Repair. V-JEPA mitigates the degradation in most cases, suggesting that the visual context encoded by Qwen3-VL and the dynamic features provided by V-JEPA offer complementary information for future location forecasting.

\begin{figure}[t]
    \centering
    \includegraphics[width=\linewidth]{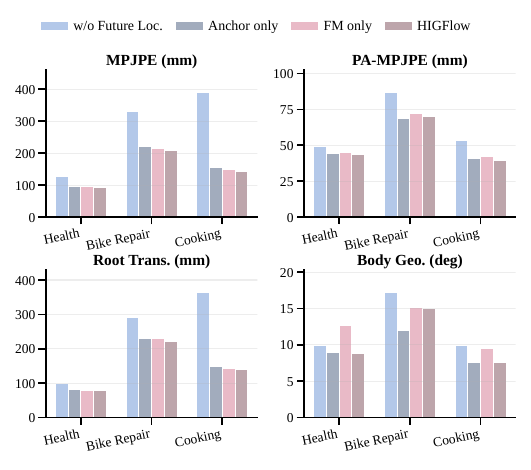}
    \caption{
    Pose stage component ablation across Health, Bike Repair, and
Cooking. We compared HIGFlow with variants without future interaction
location conditioning, without residual flow matching, and without the
deterministic anchor.
    }
    \label{fig:pose_design_ablation_all_metrics}
\end{figure}

\begin{figure}[!t]
\centering
\includegraphics[width=0.95\columnwidth]{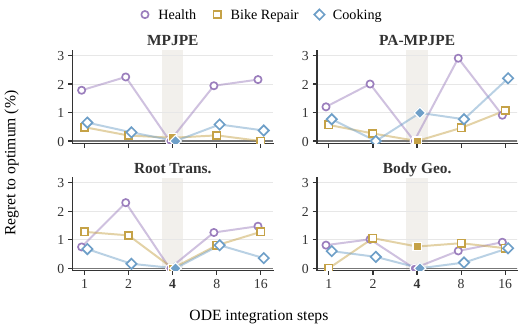}
\caption{Effect of ODE integration steps on Best-5 pose forecasting. Regret is computed relative to the best tested step count for each domain and metric. The shaded band marks the selected setting of four integration steps. Lower values are better.}
\label{fig:integration_steps_ablation}
\end{figure}

\textbf{c) V-JEPA token budget.}
We varied the number of resampled V-JEPA memory tokens over
\(M\in\{16,32,64,128\}\) while keeping all other settings fixed. Table~\ref{tab:jepa_token_budget} shows that performance does not improve monotonically with increasing memory size.
\(M=64\) achieves the best results on Bike Repair and remains competitive on Health and Cooking, providing the strongest overall trade-off across domains. We therefore adopted \(M=64\) as the default setting.

\textbf{\textit{5) How do pose stage design choices and inference settings affect pose forecasting?}}

We examined the contributions of location conditioning, the deterministic anchor, and residual flow matching, together with the effect of ODE integration steps.

\textbf{a) Location source ablation.}
The left part of Table~\ref{tab:pose_forecasting_all_combined} provides a controlled evaluation using ground-truth future interaction locations while keeping the same trained pose models, remaining inputs, and inference settings. Compared with the predicted-location setting on the right, ground-truth location conditioning generally improves pose forecasting accuracy, indicating that localization quality directly affects downstream pose prediction. HIGFlow consistently benefits from more accurate location guidance while maintaining strong overall performance under both location settings.

\textbf{b) Component ablation.}
We removed future interaction location conditioning, residual flow matching, or the deterministic anchor while keeping the observed pose history and Best-5 protocol fixed, yielding location-independent, anchor-only, and Flow-Matching-only variants, respectively.
As shown in Fig.~\ref{fig:pose_design_ablation_all_metrics}, removing future interaction location conditioning causes the largest overall degradation, demonstrating the importance of interaction locations as geometric guidance for full-body motion forecasting. The deterministic anchor improves structural stability, while residual flow matching captures additional motion variation and improves forecasting accuracy. Their combination achieves the strongest and most balanced performance across domains and metrics.

\textbf{c) Integration step analysis.}
We evaluated
\(N_{\mathrm{ODE}}\in\{1,2,4,8,16\}\)
under the same model configuration and Best-5 protocol. Because the pose metrics have different units and scales, we computed the percentage increase in error relative to the best tested result for each domain-metric pair and averaged it over all 12 pairs. 
As shown in Fig.~\ref{fig:integration_steps_ablation},
increasing the number of integration steps does not consistently reduce forecasting error. Four steps achieve the best result in 9 of the 12 comparisons and the lowest mean regret, whereas fewer steps provide insufficient integration accuracy and additional steps increase inference cost without consistent performance gains. 
We therefore used \(N_{\mathrm{ODE}}=4\) by default.

\subsection{Qualitative Analysis}

Fig.~\ref{fig:Visualization} compares HIGFlow with FIction in two
representative domains, Bike Repair and Cooking. For each example, the
upper part shows the observed pose history with synchronized exocentric
frames, while the lower part presents future reference frames together with
the ground truth and the predictions of both methods. The exocentric frames
are used only for visualization. In Bike Repair, HIGFlow preserves the bent
working pose during the early forecast and transitions to standing at a time
closer to the ground truth, whereas FIction becomes upright too early. In
Cooking, HIGFlow more often places the predicted interaction location in the
correct workspace and produces body displacement and reaching poses that are
closer to the reference motion. Together with the ablation results, these
examples support the role of future interaction locations as geometric
guidance, while the deterministic anchor and residual flow matching improve
structural stability and refine articulated motion.

The examples also reveal two limitations. In the later Bike Repair frames,
interaction location errors may grow over the forecasting horizon and
propagate through cascaded inference to the pose stage. As a result, the
predicted body remains too upright instead of following the reference
leaning and reaching motion. In Cooking, HIGFlow often identifies the correct
interaction region, but the contacting hand, contact height, arm
configuration, or torso orientation may still differ from the ground truth.
These cases suggest that interaction location conditioning improves coarse
spatial alignment but does not fully capture human intent, object
affordances, or detailed contact constraints.

\section{Conclusion}
We introduced the Coherent4D dataset, which provides temporally aligned interaction location and full-body pose sequences in a shared coordinate system for continuous supervision and evaluation. Building on this formulation, we proposed the HIGFlow framework, a cascaded where-to-how model that couples future interaction location and full-body pose forecasting by using future interaction locations as geometric conditions for pose prediction. Within HIGFlow,
Semantic-Dynamic Location Forecasting improves continuous interaction localization, while Hand-Conditioned Residual Flow Matching produces diverse yet structurally consistent poses. Experiments validate HIGFlow on both forecasting tasks, while ablations confirm the contributions of its key components. Future work will explore human intent modeling and contact-aware conditioning to enable more physically consistent forecasts over longer horizons.

\bibliographystyle{IEEEtran}
\bibliography{references}

@inproceedings{Grauman2022Ego4D,
  author    = {Grauman, Kristen and Westbury, Andrew and Byrne, Eugene and Chavis, Zachary and Furnari, Antonino and Girdhar, Rohit and Hamburger, Jackson and Jiang, Hao and Liu, Miao and Liu, Xingyu and others},
  title     = {{Ego4D}: Around the World in 3,000 Hours of Egocentric Video},
  booktitle = {Proceedings of the IEEE/CVF Conference on Computer Vision and Pattern Recognition (CVPR)},
  pages     = {18995--19012},
  year      = {2022}
}

@inproceedings{li2022egocentric,
  title={Egocentric prediction of action target in {3D}},
  author={Li, Yiming and Cao, Ziang and Liang, Andrew and Liang, Benjamin and Chen, Luoyao and Zhao, Hang and Feng, Chen},
  booktitle={2022 IEEE/CVF Conference on Computer Vision and Pattern Recognition (CVPR)},
  pages={20971--20980},
  year={2022},
  organization={IEEE}
}

@inproceedings{bao2023uncertainty,
  title={Uncertainty-aware state space transformer for egocentric {3D} hand trajectory forecasting},
  author={Bao, Wentao and Chen, Lele and Zeng, Libing and Li, Zhong and Xu, Yi and Yuan, Junsong and Kong, Yu},
  booktitle={2023 IEEE/CVF International Conference on Computer Vision (ICCV)},
  pages={13656--13665},
  year={2023},
  organization={IEEE}
}

@inproceedings{Shin2024WHAM,
  title={{WHAM}: Reconstructing world-grounded humans with accurate {3D} motion},
  author={Shin, Soyong and Kim, Juyong and Halilaj, Eni and Black, Michael J},
  booktitle={2024 IEEE/CVF Conference on Computer Vision and Pattern Recognition (CVPR)},
  pages={2070--2080},
  year={2024},
  organization={IEEE}
}

@inproceedings{ma2025diff,
  title={{Diff-IP2D}: Diffusion-based hand-object interaction prediction on egocentric videos},
  author={Ma, Junyi and Chen, Xieyuanli and Xu, Jingyi and Wang, Hesheng},
  booktitle={2025 IEEE/RSJ International Conference on Intelligent Robots and Systems (IROS)},
  pages={4291--4298},
  year={2025},
  organization={IEEE}
}

@inproceedings{ma2025novel,
  title={Novel diffusion models for multimodal {3D} hand trajectory prediction},
  author={Ma, Junyi and Bao, Wentao and Xu, Jingyi and Sun, Guanzhong and Chen, Xieyuanli and Wang, Hesheng},
  booktitle={2025 IEEE/RSJ International Conference on Intelligent Robots and Systems (IROS)},
  pages={2408--2415},
  year={2025},
  organization={IEEE}
}

@inproceedings{curreli2025nonisotropic,
  title={Nonisotropic {Gaussian} diffusion for realistic {3D} human motion prediction},
  author={Curreli, Cecilia and Muhle, Dominik and Saroha, Abhishek and Ye, Zhenzhang and Marin, Riccardo and Cremers, Daniel},
  booktitle={2025 IEEE/CVF Conference on Computer Vision and Pattern Recognition (CVPR)},
  pages={1871--1882},
  year={2025},
  organization={IEEE}
}

@inproceedings{Xu2024SLDHMP,
  title={Learning semantic latent directions for accurate and controllable human motion prediction},
  author={Xu, Guowei and Tao, Jiale and Li, Wen and Duan, Lixin},
  booktitle={European Conference on Computer Vision},
  pages={56--73},
  year={2024},
  organization={Springer}
}

@article{QwenTeam2025Qwen3VL,
  title={{Qwen3-VL} technical report},
  author={Bai, Shuai and Cai, Yuxuan and Chen, Ruizhe and Chen, Keqin and Chen, Xionghui and Cheng, Zesen and Deng, Lianghao and Ding, Wei and Gao, Chang and Ge, Chunjiang and others},
  journal={arXiv preprint arXiv:2511.21631},
  year={2025}
}

@article{liu2025sfhand,
  title={SFHand: A Streaming Framework for Language-guided 3D Hand Forecasting and Embodied Manipulation},
  author={Liu, Ruicong and Huang, Yifei and Ouyang, Liangyang and Kang, Caixin and Sato, Yoichi},
  journal={arXiv preprint arXiv:2511.18127},
  year={2025}
}

@article{chen2025flowing,
  title={Flowing from reasoning to motion: Learning {3D} hand trajectory prediction from egocentric human interaction videos},
  author={Chen, Mingfei and Wang, Yifan and Li, Zhengqin and Bharadhwaj, Homanga and Chen, Yujin and Qin, Chuan and Kou, Ziyi and Tian, Yuan and Whitmire, Eric and Sodhi, Rajinder and others},
  journal={arXiv preprint arXiv:2512.16907},
  year={2025}
}

@inproceedings{zheng2022gimo,
  title={{GIMO}: Gaze-informed human motion prediction in context},
  author={Zheng, Yang and Yang, Yanchao and Mo, Kaichun and Li, Jiaman and Yu, Tao and Liu, Yebin and Liu, C Karen and Guibas, Leonidas J},
  booktitle={European Conference on Computer Vision},
  pages={676--694},
  year={2022},
  organization={Springer}
}

@article{kratzer2020mogaze,
  title={{MoGaze}: A dataset of full-body motions that includes workspace geometry and eye-gaze},
  author={Kratzer, Philipp and Bihlmaier, Simon and Midlagajni, Niteesh Balachandra and Prakash, Rohit and Toussaint, Marc and Mainprice, Jim},
  journal={IEEE Robotics and Automation Letters},
  volume={6},
  number={2},
  pages={367--373},
  year={2020},
  publisher={IEEE}
}

@inproceedings{avogaro2024exploring,
  title={Exploring 3D human pose estimation and forecasting from the robot’s perspective: The {HARPER} dataset},
  author={Avogaro, Andrea and Toaiari, Andrea and Cunico, Federico and Xu, Xiangmin and Dafas, Haralambos and Vinciarelli, Alessandro and Li, Emma and Cristani, Marco},
  booktitle={2024 IEEE/RSJ International Conference on Intelligent Robots and Systems (IROS)},
  pages={5828--5835},
  year={2024},
  organization={IEEE}
}

@inproceedings{jiang2024map,
  title={Map-aware human pose prediction for robot follow-ahead},
  author={Jiang, Qingyuan and Susam, Burak and Chao, Jun-Jee and Isler, Volkan},
  booktitle={2024 IEEE/RSJ International Conference on Intelligent Robots and Systems (IROS)},
  pages={13031--13038},
  year={2024},
  organization={IEEE}
}

@inproceedings{fang2024egopat3dv2,
  title={{EgoPAT3Dv2}: Predicting {3D} action target from {2D} egocentric vision for human-robot interaction},
  author={Fang, Irving and Chen, Yuzhong and Wang, Yifan and Zhang, Jianghan and Zhang, Qiushi and Xu, Jiali and He, Xibo and Gao, Weibo and Su, Hao and Li, Yiming and others},
  booktitle={2024 IEEE International Conference on Robotics and Automation (ICRA)},
  pages={3036--3043},
  year={2024},
  organization={IEEE}
}

@article{Bao2025HandsOnVLM,
  title={{HandsOnVLM}: Vision-language models for hand-object interaction prediction},
  author={Bao, Chen and Xu, Jiarui and Wang, Xiaolong and Gupta, Abhinav and Bharadhwaj, Homanga},
  journal={arXiv preprint arXiv:2412.13187},
  year={2024}
}

@article{lipman2022flow,
  title={Flow matching for generative modeling},
  author={Lipman, Yaron and Chen, Ricky TQ and Ben-Hamu, Heli and Nickel, Maximilian and Le, Matt},
  journal={arXiv preprint arXiv:2210.02747},
  year={2022}
}

@article{MurLabadia2026VJEPA21,
  title={{V-JEPA 2.1}: Unlocking dense features in video self-supervised learning},
  author={Mur-Labadia, Lorenzo and Muckley, Matthew and Bar, Amir and Assran, Mido and Sinha, Koustuv and Rabbat, Mike and LeCun, Yann and Ballas, Nicolas and Bardes, Adrien},
  journal={arXiv preprint arXiv:2603.14482},
  year={2026}
}

@article{Maes2026LeWorldModel,
  title={{LeWorldModel}: Stable end-to-end joint-embedding predictive architecture from pixels},
  author={Maes, Lucas and Lidec, Quentin Le and Scieur, Damien and LeCun, Yann and Balestriero, Randall},
  journal={arXiv preprint arXiv:2603.19312},
  year={2026}
}

@inproceedings{liu2022joint,
  title={Joint hand motion and interaction hotspots prediction from egocentric videos},
  author={Liu, Shaowei and Tripathi, Subarna and Majumdar, Somdeb and Wang, Xiaolong},
  booktitle={2022 IEEE/CVF Conference on Computer Vision and Pattern Recognition (CVPR)},
  pages={3272--3282},
  year={2022},
  organization={IEEE}
}

@article{ma2026madiff,
  title={{MADiff}: Motion-Aware Mamba Diffusion Models for Hand Trajectory Prediction on Egocentric Videos},
  author={Ma, Junyi and Chen, Xieyuanli and Bao, Wentao and Xu, Jingyi and Wang, Hesheng},
  journal={IEEE Transactions on Pattern Analysis and Machine Intelligence},
  volume={48},
  number={3},
  pages={3250--3267},
  year={2026},
  publisher={IEEE}
}

@article{hatano2025invisible,
  title={The invisible {EgoHand}: {3D} hand forecasting through egobody pose estimation},
  author={Hatano, Masashi and Zhu, Zhifan and Saito, Hideo and Damen, Dima},
  journal={arXiv preprint arXiv:2504.08654},
  year={2025}
}

@article{ma2026uni,
  title={{Uni-Hand}: Universal Hand Motion Forecasting in Egocentric Views},
  author={Ma, Junyi and Bao, Wentao and Xu, Jingyi and Sun, Guanzhong and Zheng, Yu and Zhang, Erhang and Chen, Xieyuanli and Wang, Hesheng},
  journal={IEEE Transactions on Pattern Analysis and Machine Intelligence},
  year={2026},
  note={{Early Access}},
  publisher={IEEE}
}

@inproceedings{Xu2022STARS,
  title={Diverse human motion prediction guided by multi-level spatial-temporal anchors},
  author={Xu, Sirui and Wang, Yu-Xiong and Gui, Liang-Yan},
  booktitle={European Conference on Computer Vision},
  pages={251--269},
  year={2022},
  organization={Springer}
}

@inproceedings{wei2023human,
  title={Human joint kinematics diffusion-refinement for stochastic motion prediction},
  author={Wei, Dong and Sun, Huaijiang and Li, Bin and Lu, Jianfeng and Li, Weiqing and Sun, Xiaoning and Hu, Shengxiang},
  booktitle={Proceedings of the AAAI Conference on Artificial Intelligence},
  volume={37},
  number={5},
  pages={6110--6118},
  year={2023}
}

@inproceedings{chen2023humanmac,
  title={{HumanMAC}: Masked motion completion for human motion prediction},
  author={Chen, Ling-Hao and Zhang, Jiawei and Li, Yewen and Pang, Yiren and Xia, Xiaobo and Liu, Tongliang},
  booktitle={2023 IEEE/CVF International Conference on Computer Vision (ICCV)},
  pages={9510--9521},
  year={2023},
  organization={IEEE}
}

@inproceedings{Sun2024CoMusion,
  title={{CoMusion}: Towards consistent stochastic human motion prediction via motion diffusion},
  author={Sun, Jiarui and Chowdhary, Girish},
  booktitle={European conference on computer vision},
  pages={18--36},
  year={2024},
  organization={Springer}
}

@inproceedings{Jeong2024T2P,
  title={Multi-agent long-term {3D} human pose forecasting via interaction-aware trajectory conditioning},
  author={Jeong, Jaewoo and Park, Daehee and Yoon, Kuk-Jin},
  booktitle={2024 IEEE/CVF Conference on Computer Vision and Pattern Recognition (CVPR)},
  pages={16975--16984},
  year={2024},
  organization={IEEE}
}

@inproceedings{Yu2025GAP3DS,
  title={Vision-guided action: Enhancing {3D} human motion prediction with gaze-informed affordance in {3D} scenes},
  author={Yu, Ting and Lin, Yi and Yu, Jun and Lou, Zhenyu and Cui, Qiongjie},
  booktitle={2025 IEEE/CVF Conference on Computer Vision and Pattern Recognition (CVPR)},
  pages={12335--12346},
  year={2025},
  organization={IEEE}
}

@article{Tian2025PrediFlow,
  title={{PrediFlow}: A Flow-Based Prediction-Refinement Framework for Real-Time Human Motion Prediction in Human-Robot Collaboration},
  author={Tian, Sibo and Zheng, Minghui and Liang, Xiao},
  journal={arXiv preprint arXiv:2512.13903},
  year={2025}
}

@inproceedings{li2025satori,
  title={Satori: Towards Proactive AR Assistant with Belief-Desire-Intention User Modeling},
  author={Li, Chenyi and Wu, Guande and Chan, Gromit Yeuk-Yin and Turakhia, Dishita Gdi and Castelo Quispe, Sonia and Li, Dong and Welch, Leslie and Silva, Claudio and Qian, Jing},
  booktitle={Proceedings of the 2025 CHI Conference on Human Factors in Computing Systems},
  pages={1--24},
  year={2025}
}

@inproceedings{pei2025attentionar,
  title={{AttentionAR}: Ar adaptation and warning for real-world safety via attention modeling and mllm reasoning},
  author={Pei, Yunqiang and Huang, Renming and Zha, Mingfeng and Wang, Guoqing and Wang, Peng and Kang, Qiao and Yang, Yang and Shen, Heng Tao},
  booktitle={Proceedings of the 38th Annual ACM Symposium on User Interface Software and Technology},
  pages={1--19},
  year={2025}
}

@article{noormohammadi2025lead,
  title={To lead or to follow? Adaptive robot task planning in human--robot collaboration},
  author={Noormohammadi-Asl, Ali and Smith, Stephen L and Dautenhahn, Kerstin},
  journal={IEEE Transactions on Robotics},
  volume={41},
  pages={4215--4235},
  year={2025},
  publisher={IEEE}
}

@inproceedings{barquero2023belfusion,
  title={{BeLFusion}: Latent diffusion for behavior-driven human motion prediction},
  author={Barquero, German and Escalera, Sergio and Palmero, Cristina},
  booktitle={2023 IEEE/CVF International Conference on Computer Vision (ICCV)},
  pages={2317--2327},
  year={2023},
  organization={IEEE}
}

@inproceedings{Grauman2024EgoExo4D,
  title={{Ego-Exo4D}: Understanding skilled human activity from first -and third-person perspectives},
  author={Grauman, Kristen and Westbury, Andrew and Torresani, Lorenzo and Kitani, Kris and Malik, Jitendra and Afouras, Triantafyllos and Ashutosh, Kumar and Baiyya, Vijay and Bansal, Siddhant and Boote, Bikram and others},
  booktitle={Proceedings of the IEEE/CVF conference on computer vision and pattern recognition},
  pages={19383--19400},
  year={2024}
}

@inproceedings{ashutosh2025fiction,
  title={{FICTION}: {4D} future interaction prediction from video},
  author={Ashutosh, Kumar and Pavlakos, Georgios and Grauman, Kristen},
  booktitle={2025 IEEE/CVF Conference on Computer Vision and Pattern Recognition (CVPR)},
  pages={17613--17625},
  year={2025},
  organization={IEEE}
}

@inproceedings{zhou2022detecting,
  title={Detecting twenty-thousand classes using image-level supervision},
  author={Zhou, Xingyi and Girdhar, Rohit and Joulin, Armand and Kr{\"a}henb{\"u}hl, Philipp and Misra, Ishan},
  booktitle={European conference on computer vision},
  pages={350--368},
  year={2022},
  organization={Springer}
}

@inproceedings{Gupta2019LVIS,
  title={{LVIS}: A dataset for large vocabulary instance segmentation},
  author={Gupta, Agrim and Dollar, Piotr and Girshick, Ross},
  booktitle={2019 IEEE/CVF Conference on Computer Vision and Pattern Recognition (CVPR)},
  pages={5351--5359},
  year={2019},
  organization={IEEE}
}

@article{grattafiori2024llama3,
  title={The {Llama 3} herd of models},
  author={Grattafiori, Aaron and Dubey, Abhimanyu and Jauhri, Abhinav and Pandey, Abhinav and Kadian, Abhishek and Al-Dahle, Ahmad and Letman, Aiesha and Mathur, Akhil and Schelten, Alan and Vaughan, Alex and others},
  journal={arXiv preprint arXiv:2407.21783},
  year={2024}
}

@incollection{loper2023smpl,
  title={{SMPL}: A skinned multi-person linear model},
  author={Loper, Matthew and Mahmood, Naureen and Romero, Javier and Pons-Moll, Gerard and Black, Michael J},
  booktitle={Seminal Graphics Papers: Pushing the Boundaries, Volume 2},
  pages={851--866},
  year={2023}
}

@article{seminara2026task,
  title={Task Graph Maximum Likelihood Estimation for Procedural Activity Understanding in Egocentric Videos},
  author={Seminara, Luigi and Farinella, Giovanni Maria and Furnari, Antonino},
  journal={IEEE Transactions on Pattern Analysis and Machine Intelligence},
  volume={48},
  number={9},
  pages={10518--10534},
  year={2026},
  publisher={IEEE}
}

@article{liu2026goal,
  title={Goal-Guided Prompting With Adaptive Modality Selection for Efficient Assembly Activity Anticipation in Egocentric Videos},
  author={Liu, Tianshan and Bao, Bing-Kun},
  journal={IEEE Transactions on Pattern Analysis and Machine Intelligence},
  volume={48},
  number={5},
  pages={5945--5962},
  year={2026},
  publisher={IEEE}
}

@article{liu2022investigating,
  title={Investigating pose representations and motion contexts modeling for 3D motion prediction},
  author={Liu, Zhenguang and Wu, Shuang and Jin, Shuyuan and Ji, Shouling and Liu, Qi and Lu, Shijian and Cheng, Li},
  journal={IEEE Transactions on Pattern Analysis and Machine Intelligence},
  volume={45},
  number={1},
  pages={681--697},
  year={2022},
  publisher={IEEE}
}

@article{shu2021spatiotemporal,
  title={Spatiotemporal co-attention recurrent neural networks for human-skeleton motion prediction},
  author={Shu, Xiangbo and Zhang, Liyan and Qi, Guo-Jun and Liu, Wei and Tang, Jinhui},
  journal={IEEE Transactions on Pattern Analysis and Machine Intelligence},
  volume={44},
  number={6},
  pages={3300--3315},
  year={2021},
  publisher={IEEE}
}

@article{peirone2026hier,
  title={Hier-EgoPack: Hierarchical Egocentric Video Understanding With Diverse Task Perspectives},
  author={Peirone, Simone Alberto and Pistilli, Francesca and Alliegro, Antonio and Tommasi, Tatiana and Averta, Giuseppe},
  journal={IEEE Transactions on Pattern Analysis and Machine Intelligence},
  volume={48},
  number={2},
  pages={1917--1931},
  year={2026},
  publisher={IEEE}
}

@article{mur2026integrating,
  title={Integrating Affordances and Attention Models for Short-Term Object Interaction Anticipation},
  author={Mur-Labadia, Lorenzo and Martinez-Cantin, Ruben and Guerrero, Jose J and Farinella, Giovanni Maria and Furnari, Antonino},
  journal={IEEE Transactions on Pattern Analysis and Machine Intelligence},
  volume={48},
  number={5},
  pages={5425--5441},
  year={2026},
  publisher={IEEE}
}

@article{zeng2023x,
  title={{X$^2$-VLM}: All-in-One Pre-Trained Model for Vision-Language Tasks},
  author={Zeng, Yan and Zhang, Xinsong and Li, Hang and Wang, Jiawei and Zhang, Jipeng and Zhou, Wangchunshu},
  journal={IEEE Transactions on Pattern Analysis and Machine Intelligence},
  volume={46},
  number={5},
  pages={3156--3168},
  year={2023},
  publisher={IEEE}
}

@article{tian2026ego,
  title={Ego-R1: Agentic Chain-of-Tool-Thought for Ultra-Long Egocentric Video Reasoning},
  author={Tian, Shulin and Wang, Ruiqi and Guo, Hongming and Wu, Penghao and Dong, Yuhao and Wang, Xiuying and Yang, Jingkang and Zhang, Hao and Zhu, Hongyuan and Liu, Ziwei},
  journal={IEEE Transactions on Pattern Analysis and Machine Intelligence},
  year={2026},
  note={{Early Access}},
  publisher={IEEE}
}

@article{Chen2025MotionLLM,
  author  = {Chen, Ling-Hao and Lu, Shunlin and Zeng, Ailing and Zhang, Hao and Wang, Benyou and Zhang, Ruimao and Zhang, Lei},
  title   = {{MotionLLM}: Understanding Human Behaviors from Human Motions and Videos},
  journal = {IEEE Transactions on Pattern Analysis and Machine Intelligence},
  year    = {2025},
  note    = {{Early Access}},
  doi     = {10.1109/TPAMI.2025.3627546}
}

@article{Fernando2025Remembering,
  title={Remembering what is important: a factorised multi-head retrieval and auxiliary memory stabilisation scheme for human motion prediction},
  author={Fernando, Tharindu and Gammulle, Harshala and Sridharan, Sridha and Denman, Simon and Fookes, Clinton},
  journal={IEEE Transactions on Pattern Analysis and Machine Intelligence},
  volume={47},
  number={3},
  pages={1941--1957},
  year={2024},
  publisher={IEEE}
}

@article{Tang2026ContinualPrior,
  title={Human Motion Prediction via Continual Prior Compensation},
  author={Tang, Jianwei and Hu, Jian-Fang and Liang, Tianming and Lin, Xiaotong and Sun, Jiangxin and Zheng, Wei-Shi and Lai, Jianhuang},
  journal={IEEE Transactions on Pattern Analysis and Machine Intelligence},
  volume={48},
  number={5},
  pages={5131--5146},
  year={2026},
  publisher={IEEE}
}

@article{Zhu2024HumanMotionSurvey,
  title={Human motion generation: A survey},
  author={Zhu, Wentao and Ma, Xiaoxuan and Ro, Dongwoo and Ci, Hai and Zhang, Jinlu and Shi, Jiaxin and Gao, Feng and Tian, Qi and Wang, Yizhou},
  journal={IEEE Transactions on Pattern Analysis and Machine Intelligence},
  volume={46},
  number={4},
  pages={2430--2449},
  year={2023},
  publisher={IEEE}
}

@article{Zhang2024MotionDiffuse,
  title={{MotionDiffuse}: Text-driven human motion generation with diffusion model},
  author={Zhang, Mingyuan and Cai, Zhongang and Pan, Liang and Hong, Fangzhou and Guo, Xinying and Yang, Lei and Liu, Ziwei},
  journal={IEEE Transactions on Pattern Analysis and Machine Intelligence},
  volume={46},
  number={6},
  pages={4115--4128},
  year={2024},
  publisher={IEEE}
}

@article{Yang2026LagrangianMotionFields,
  title={Lagrangian Motion Fields for Long-Term Motion Generation},
  author={Yang, Yifei and Huang, Zikai and Xu, Chenshu and He, Shengfeng},
  journal={IEEE Transactions on Pattern Analysis and Machine Intelligence},
  volume={48},
  number={2},
  pages={1171--1184},
  year={2026},
  publisher={IEEE}
}

@article{damen2020epic,
  title={The {EPIC-KITCHENS} dataset: Collection, challenges and baselines},
  author={Damen, Dima and Doughty, Hazel and Farinella, Giovanni Maria and Fidler, Sanja and Furnari, Antonino and Kazakos, Evangelos and Moltisanti, Davide and Munro, Jonathan and Perrett, Toby and Price, Will and others},
  journal={IEEE Transactions on Pattern Analysis and Machine Intelligence},
  volume={43},
  number={11},
  pages={4125--4141},
  year={2020},
  publisher={IEEE}
}

@article{Liu2026STKAD,
  title={Anticipating Object Interactions Via Aggregation and Distillation of Spatio-Temporal Knowledge From Vision Language Models},
  author={Liu, Yang and Yang, Dejie and Zheng, Minghang and Yang, Ming-Hsuan},
  journal={IEEE Transactions on Pattern Analysis and Machine Intelligence},
  year={2026},
  note={{Early Access}},
  doi={10.1109/TPAMI.2026.3726073},
  publisher={IEEE}
}

@article{qi2024uncertainty,
  title={Uncertainty-Boosted Robust Video Activity Anticipation},
  author={Qi, Zhaobo and Wang, Shuhui and Zhang, Weigang and Huang, Qingming},
  journal={IEEE Transactions on Pattern Analysis and Machine Intelligence},
  volume={46},
  number={12},
  pages={7775--7792},
  year={2024},
  publisher={IEEE}
}

@article{Chang2025STAU,
  title={STAU: a spatiotemporal-aware unit for video prediction and beyond},
  author={Chang, Zheng and Zhang, Xinfeng and Wang, Shanshe and Ma, Siwei and Gao, Wen},
  journal={IEEE Transactions on Pattern Analysis and Machine Intelligence},
  volume={47},
  number={9},
  pages={7916--7929},
  year={2025},
  publisher={IEEE}
}

\end{document}